\documentclass[]{fairmeta}
\usepackage{enumitem}
\usepackage{tabularx} 
\usepackage{listings}
\usepackage{multirow}
\usepackage{makecell}
\usepackage{graphicx}
\usepackage{wrapfig}
\usepackage{tabularx}
\usepackage{textcomp}
\usepackage{stfloats}
\usepackage{url}
\usepackage{verbatim}
\usepackage{titlesec}
\usepackage{tocloft}
\usepackage{adjustbox}
\usepackage{multirow}
\usepackage{pifont}
\usepackage{tikz}
\usepackage{comment}
\usepackage{amsmath,amssymb} 
\usepackage{colortbl}  
\usepackage{color}
\usepackage{booktabs} 
\usepackage{hyperref}
\usepackage{graphicx}    
\usepackage{subcaption} 
\usepackage{multirow} 
\usepackage{booktabs} 
\usepackage{subcaption} 
\RequirePackage{xspace}
\makeatletter
\DeclareRobustCommand\onedot{\futurelet\@let@token\@onedot}
\def\@onedot{\ifx\@let@token.\else.\null\fi\xspace}
\usepackage[most]{tcolorbox}
\usepackage{xcolor}
\setcitestyle{square,comma,numbers,sort&compress}

\makeatother

\definecolor{adptorange}{RGB}{248, 205, 172}
\definecolor{cmpblue}{RGB}{189, 215, 238}
\definecolor{cmpblue}{RGB}{189, 215, 238}

\definecolor{our_red}{RGB}{232,157,160}
\definecolor{our_blue}{RGB}{136,206,230}
\definecolor{our_orange}{RGB}{246,200,168}
\definecolor{our_green}{RGB}{178,211,164}

\definecolor{attn_code0}{RGB}{247,215,200}
\definecolor{attn_code1}{RGB}{238,169,139}
\definecolor{mlp_code0}{RGB}{204,201,221}
\definecolor{mlp_code1}{RGB}{102,95,153}

\definecolor{token_blue}{RGB}{84, 120, 140}

\definecolor{myblue}{RGB}{233, 241, 249}
\definecolor{mygray}{RGB}{99, 110, 114}
\definecolor{myred}{RGB}{255, 118, 117}
\definecolor{myyellow}{RGB}{255, 234, 167}
\definecolor{mygreen}{RGB}{216, 226, 204}
\definecolor{mypurple}{RGB}{162, 155, 254}
\definecolor{mybrown}{RGB}{215, 190, 154}
\definecolor{myorange}{RGB}{255, 220, 190}

\usepackage{pifont}       
\usepackage{bbding}       
\usepackage{fontawesome}
\usepackage{xspace}

\usepackage{float}

\newlength\savewidth

\newcolumntype{x}[1]{>{\centering\arraybackslash}p{#1pt}}
\newcolumntype{y}[1]{>{\raggedright\arraybackslash}p{#1pt}}
\newcolumntype{z}[1]{>{\raggedleft\arraybackslash}p{#1pt}}

\renewcommand{\paragraph}[1]{\vspace{1mm}\noindent\textbf{#1}}
\usepackage{colortbl}
\usepackage{xcolor}
\usepackage{wrapfig}

\renewcommand{\paragraph}[1]{\vspace{1.25mm}\noindent\textbf{#1}}

\usepackage{algorithm}
\usepackage{listings}

\definecolor{codeblue}{rgb}{0.25, 0.5, 0.5}
\definecolor{codekw}{rgb}{0.35, 0.35, 0.75}
\lstdefinestyle{Pytorch}{
    language = Python,
    backgroundcolor = \color{white},
    basicstyle = \fontsize{9pt}{8pt}\selectfont\ttfamily\bfseries,
    columns = fullflexible,
    aboveskip=1pt,
    belowskip=1pt,
    breaklines = true,
    captionpos = b,
    commentstyle = \color{codeblue},
    keywordstyle = \color{codekw},
}

\definecolor{green}{HTML}{009000}
\definecolor{red}{HTML}{ea4335}

\title{RynnEC: Bringing MLLMs into Embodied World}

\author[* 1, 2]{Ronghao Dang}
\author[* 1, 3]{Yuqian Yuan}
\author[* 1, 3]{Yunxuan Mao}
\author[* 1, 2]{Kehan Li}
\author[1, 3]{Jiangpin Liu}
\author[1, 2]{Zhikai Wang}
\author[1]{\\  Fan Wang}
\author[1, 2]{Deli Zhao}
\author[1, 2]{Xin Li}

\affiliation[1]{DAMO Academy, Alibaba Group\\}
\affiliation[2]{Hupan Lab}
\affiliation[3]{Zhejiang University}
\contribution[*]{Equal contribution}

\abstract{

We introduce \textbf{RynnEC}, a video multimodal large language model designed for embodied cognition. Built upon a general-purpose vision-language foundation model, RynnEC incorporates a region encoder and a mask decoder, enabling flexible region-level video interaction. Despite its compact architecture, RynnEC achieves state-of-the-art performance in object property understanding, object segmentation, and spatial reasoning. 
Conceptually, it offers a region-centric video paradigm for the brain of embodied agents, providing fine-grained perception of the physical world and enabling more precise interactions.
To mitigate the scarcity of annotated 3D datasets, we propose an egocentric video based pipeline for generating embodied cognition data. Furthermore, we introduce \textbf{RynnEC-Bench}, a region-centered benchmark for evaluating embodied cognitive capabilities. We anticipate that RynnEC will advance the development of general-purpose cognitive cores for embodied agents and facilitate generalization across diverse embodied tasks.
The code, model checkpoints, and benchmark are available at: \url{https://github.com/alibaba-damo-academy/RynnEC}

}

\date{\today}

\begin{document}
\thispagestyle{firstheader}
\maketitle
\pagestyle{empty}

\begin{figure*}[ht]
\includegraphics[width=\textwidth]{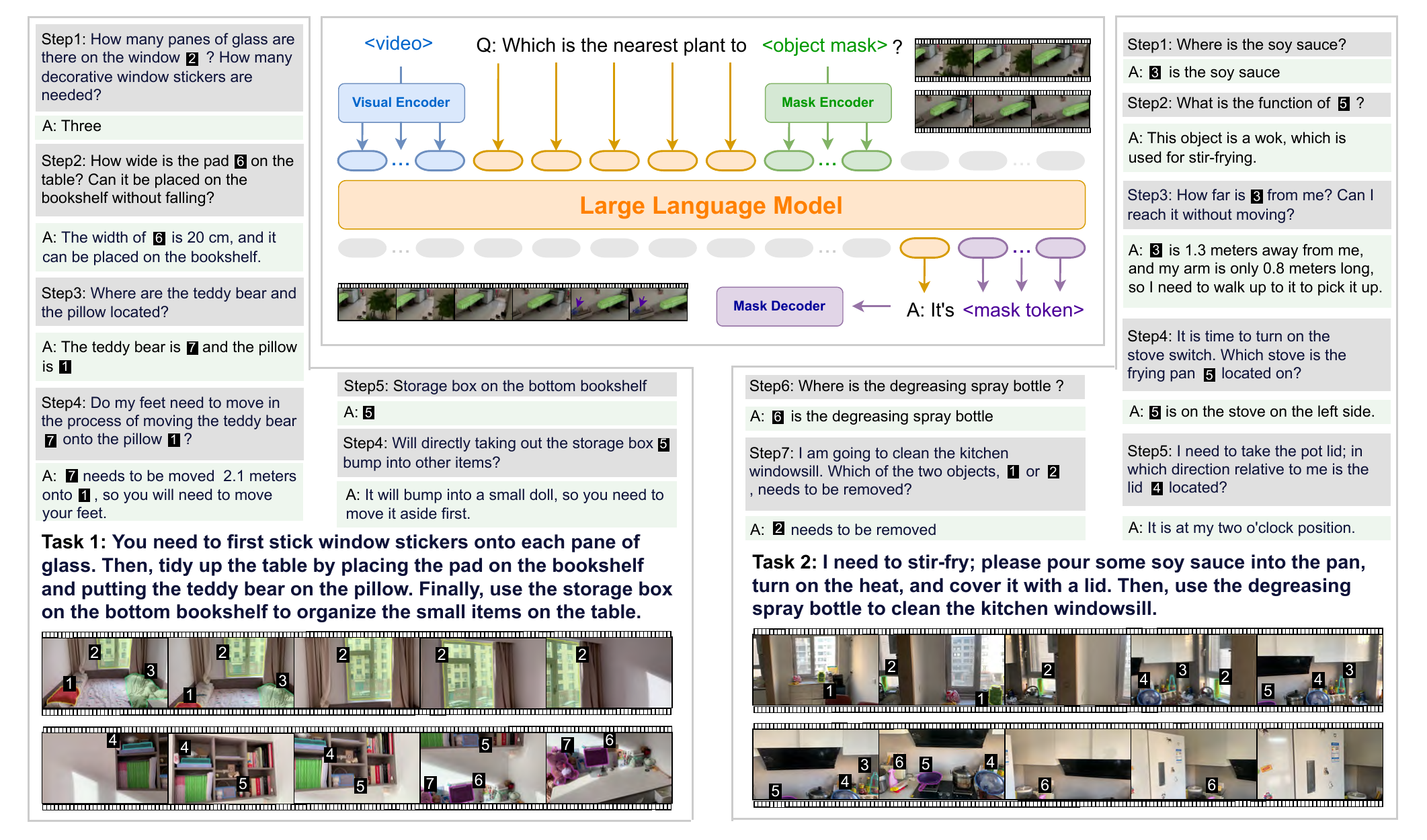}
\caption{\textbf{RynnEC} is a video multi-modal large language model (MLLM) specifically designed for embodied cognition tasks. It can accept inputs interwoven from video, region masks, and text, and produce output in the form of text or masks based on the question. RynnEC is capable of addressing a diverse range of object and spatial questions within embodied contexts and plays a significant role in indoor embodied tasks.} 
\label{fig:intro}
\end{figure*}

\section{Introduction} \label{sec:introduction}
\noindent 




In recent years, Multi-modal Large Language Models (MLLMs)~\cite{mllm_survey, mm-llms} have experienced rapid development, leading to the emergence of models such as Gemini~\cite{gemini} and GPT-4o~\cite{gpt4o} that can handle image and even video inputs. These MLLMs are attracting increasing attention from researchers due to their powerful contextual understanding~\cite{icl_mllm} and generalization~\cite{mllm_general} abilities. Researchers in embodied intelligence are also beginning to explore the use of MLLMs as the brains of robots~\cite{vlm_robot, mllm_grasp}, enabling them to perceive the real world through visual inputs like humans do. However, the current mainstream MLLMs are trained on extensive internet images and lack the foundational visual cognition to match the physical world~\cite{ecbench, eocbench}.


Some works have begun exploring how MLLMs can be applied to ego-centric embodied scenarios. Models like Exo2Ego~\cite{exo2ego} and EgoLM~\cite{egolm} enhance the understanding of ego-centric dynamic environment interactions. SpatialVLM~\cite{Spatialvlm} and SpatialRGPT~\cite{spatialrgpt} focus on addressing spatial understanding challenges within embodied contexts. 
However, these approaches are challenging to directly implement in physical robots to perform complex tasks. The main limitations are as follows:

\begin{enumerate}
    \item \textbf{Lack of flexible visual interaction:} In complex embodied scenarios, relying solely on textual communication is prone to ambiguity or vagueness. Direct visual interaction references, such as masks or points, can more accurately and flexibly index entities within a scene, facilitating precise task execution.

    \item \textbf{Insufficient detailed understanding of objects:} During task execution, objects typically serve as the smallest operational units, making comprehensive and detailed understanding of objects crucial. As illustrated in \textit{Task 1 Step 1} in Fig.~\ref{fig:intro}, recognizing the number of panes in a window is essential to determine the quantity of window decals needed.

    \item \textbf{Absence of video-based coherent spatial awareness:} For humans, spatial cognition arises from continuous visual perception~\cite{visual_SC}. Current methods in spatial intelligence~\cite{SPAR, Multi-spatialmllm} primarily focus on single or discrete images, lacking the capacity for spatial understanding in high-continuity videos. For example, in \textit{Task 1 Step 4} in Fig.~\ref{fig:intro}, the absolute distance between the teddy bear and the pillow requires a spatial scale concept derived from the entire video to be properly inferred.
\end{enumerate}


Thus, we propose \textbf{RynnEC}, an embodied cognitive MLLM designed to enhance robotic understanding of the physical world. As illustrated in Fig.~\ref{fig:intro}, RynnEC is a large video understanding model whose visual encoder and foundational parameters are derived from VideoLLaMA3~\cite{videollama3}. To enable flexible visual interaction, we incorporate an encoder and decoder specifically for region masks in videos, allowing RynnEC to achieve precise instance-level comprehension and grounding.


Within this framework, RynnEC is designed to perform diverse cognitive tasks in embodied scenarios. We categorize embodied cognitive abilities into two essential components: object cognition and spatial cognition. Object cognition necessitates MLLMs' understanding of object attributes, quantities, and their relationships with the environment, alongside accurate object grounding. Spatial cognition is further divided into world-centric and ego-centric perspectives. World-centric spatial cognition requires the model to grasp absolute scales and relative positions within scenes, as exemplified by object size estimations in \textit{Task 1 Step 2} (Fig.~\ref{fig:intro}). Ego-centric spatial cognition connects the robot's physical embodiment with the world, thereby assisting in behavioral decisions. For example, as depicted in Fig.~\ref{fig:intro}, the reachability estimation in \textit{Task 2 Step 3} and the orientation estimation in \textit{Task 2 Step 5} assist the robot in clearly defining its relationship with interactive objects.
Equipped with enhanced object and spatial reasoning, RynnEC supports more efficient execution of complex, real-world robotic tasks.


Regrettably, the development of embodied cognition models has been slow due to a lack of ego-centric videos and high-quality annotations. Efforts such as Multi-SpatialMLLM~\cite{Multi-spatialmllm}, Spatial-MLLM~\cite{Spatial-MLLM}, and SpaceR~\cite{SpaceR} leverage open-source datasets with comprehensive 3D point cloud and annotations to generate training data. However, in an era of scarce 3D annotations~\cite{bloomscene, mmscan}, this approach cannot achieve rapid and cost-effective expansion of data scale. Hence, we propose a data generation pipeline that transforms ego-centric RGB videos into embodied cognition question-answering datasets. This pipeline begins with instance segmentation from videos and diverges into two branches: one generating object cognition data and the other producing spatial cognition data. Ultimately, data from both branches are integrated into a comprehensive embodied cognition dataset. 
From over 200 households, we collect more than 20,000 egocentric videos. A subset from ten households is manually verified and balanced to create \textbf{RynnEC-Bench}, a fine-grained embodied cognition benchmark encompassing 22 tasks in object and spatial cognition.


Extensive experiments demonstrate that RynnEC significantly outperforms both general~\cite{gpt4o, qwen25vl, InternVL3} and task-specific~\cite{sa2va, videorefer, robobrain2.0} MLLMs in cognitive abilities within embodied scenarios, showcasing scalable application potential. Additionally, we observe notable advantages in multi-task training with RynnEC and identify preliminary signs of emergence in more challenging embodied cognition tasks. Finally, we highlight the potential of RynnEC in facilitating robots to undertake large-scale, long-range tasks.

\section{Related Work}
\noindent 
\subsection{MLLMs for Video Understanding}

Early MLLMs primarily relied on sparse sampling and simple connectors, such as MLPs~\cite{Video-llava, Minigpt4-video, Video-chatgpt} and Q-Formers~\cite{Video-llama, mvbench}, to integrate visual representation with large language models. Subsequently, to tackle the problem of long video understanding, \cite{long_context} directly expanded the context window of language models, while \cite{Video_instruction_tuning} introduced pooling in the spatial and temporal dimensions to compress the number of video tokens. As the need for more fine-grained understanding emerged, some studies (VideoRefer~\cite{videorefer}, DAM~\cite{dam} and PAM~\cite{PAM}) employed region-level feature encoders enabling video MLLMs to accept masked inputs and comprehend the semantic features of objects within the masks. Although these video MLLMs have demonstrated superior capabilities in high-level semantic capture and temporal modeling, they lack robust physical-world comprehension in egocentric embodied scenarios.


\subsection{Embodied Scene Understanding Benchmarks}

Some studies~\cite{explore_eqa, manipllm, MLLM_ROBOT} have begun to explore leveraging MLLMs to assist robots in solving embodied tasks. However, determining whether these MLLMs possess the ability to understand and interact with the physical world is challenging. Consequently, several benchmarks have emerged to evaluate the capability of MLLMs to perceive the physical world. OpenEQA~\cite{openeqa} and IndustryEQA~\cite{industryeqa} focus on several key competencies in home and industrial settings, respectively, and manually designed open-vocabulary questions. VSI-Bench~\cite{vsibench} centers on assessing the spatial cognitive abilities of MLLMs. STI-Bench~\cite{stibench} introduces more complex kinematic (e.g. velocity) problems. ECBench~\cite{ecbench} systematically categorizes embodied cognitive abilities into static environments, dynamic environments, and overcoming hallucinations, offering a comprehensive evaluation across 30 sub-competencies. While these benchmarks encompass a wide range of abilities, they are unable to assess more fine-grained, region-level understanding capabilities in embodied scenarios. Compared to purely textual question-answering, region-level visual interaction can more accurately refer to targets in the complex real world.

\subsection{Improving MLLMs for Embodied Cognition}

The aforementioned embodied benchmarks have highlighted the cognitive limitations of current MLLMs in embodied scenarios. Consequently, some studies have started to investigate diverse strategies for enhancing MLLMs’ understanding of the physical world. GPT4Scene~\cite{GPT4Scene} improves MLLMs' consistent global scene understanding by explicitly adding instance marks between video frames. SAT~\cite{sat} explores multi-frame dynamic spatial reasoning in simulated environments. Spatial-MLLM~\cite{Spatial-MLLM}, Multi-SpatialMLLM~\cite{Multi-spatialmllm}, and SpaceR~\cite{SpaceR} leverage 3D datasets with detailed annotations (e.g., ScanNet~\cite{scannet++}) to construct the suite of spatial-intelligence tasks introduced in VSI-Bench. In contrast, our data generation pipeline based on RGB videos yields more realistic and scalable training data. 
More importantly, RynnEC is designed not just to handle selected capabilities in embodied scenarios, but to cover a broad swath of the world cognition required for embodied task execution under a single paradigm.

\section{Methodology}
\noindent

RynnEC is a robust 
video embodied cognition model capable of processing and outputting various video object proposals. This enables it to flexibly address embodied questions about objects and space. Due to a paucity of research in this domain, we comprehensively present the construction process of RynnEC from four perspectives: data generation (Sec.~\ref{sec:data_generation}), evaluation framework establishment (Sec.~\ref{sec:RynnEC-Bench}), model architecture (Sec.~\ref{sec:archi}), and training (Sec.~\ref{sec:training}).

\begin{figure*}[t]
\includegraphics[width=\textwidth]{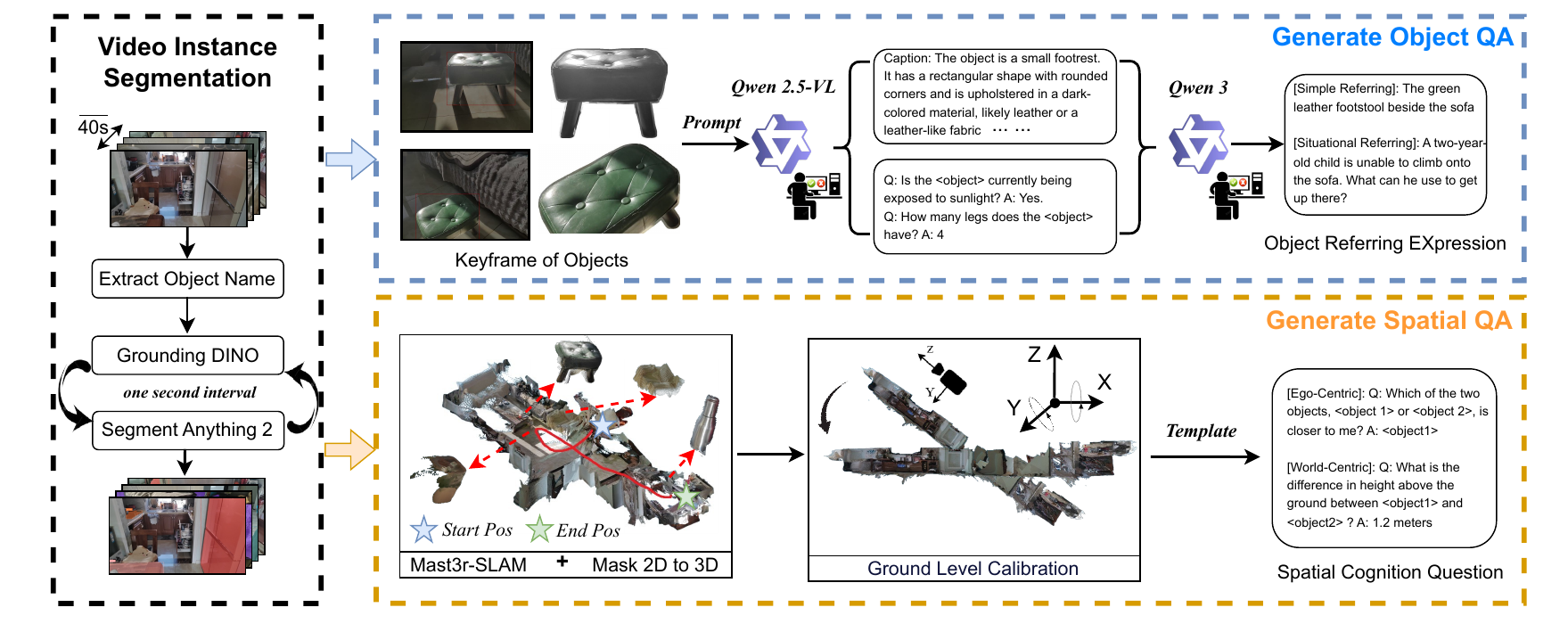}
\caption{
\textbf{Embodied Cognition Question-Answer (QA) Data Generation Pipeline:} 
First, objects within the scene are segmented from the video. Subsequently, object and spatial QA pairs are generated via two distinct branches.} 
\label{fig:data_pipeline}
\end{figure*}

\subsection{Embodied Cognition Data Generation}
\label{sec:data_generation}

Our embodied cognition dataset construction (Fig.~\ref{fig:data_pipeline}) begins with egocentric video collection and instance segmentation. One branch employs a human-in-the-loop streaming generation approach to construct various object cognition QA pairs. The other branch utilizes a monocular dense 3D reconstruction method and diverse question templates to generate spatial cognition task QA pairs.

\subsubsection{Video Collection and Instance Segmentation}
\label{sec:instance_seg}

Our egocentric video collection encompasses $200+$ houses, with approximately 100 videos per house. To ensure video quality, we require a resolution of at least 1080p and a frame rate no less than 30fps, using a gimbal to maintain shooting stability. To achieve diversity among different video trajectories, each house is divided into multiple zones, with filming trajectories categorized into single-zone, dual-zone, and tri-zone types. Cross-zone filming enhances diversity by altering the sequence of traversed zones. Additionally, we randomly vary lighting conditions and camera height under different trajectories. We require that each video includes both vertical and horizontal rotations, as well as at least two close-ups of objects, simulating the variable field of view in robotic task execution. Ultimately, we collect 20,832 egocentric videos of indoor movement. To control video length, these videos are segmented every 40 seconds.


Previous works~\cite{VeBrain, embodiedscan} adopted a strategy of designing separate data generation processes for each task type, leading to limited data reusability and continuity. We aim to create a lineage among different types of foundational data to reduce unnecessary redundancy in data generation. Therefore, this paper proposes a mask-centric embodied cognition QA generation pipeline. This pipeline initiates with the generation of object masks from video instance segmentation within a scene. First, Qwen2.5-VL~\cite{qwen25vl} observes the raw video and outputs an object list containing the names of all entity categories in the scene. Utilizing this object list, Grounding DINO 1.5~\cite{Grounding_dino_1.5} detects objects in key frames at one-second intervals. SAM2~\cite{sam2} assists in segmenting and tracking the objects detected by Grounding DINO 1.5 during the intervening one-second interval. To ensure consistency of instance IDs, the tracking results of old instances are compared with the segmentation results of newly detected instances at key frames. If an instance is found to have overlapping masks (IOU > 0.5), it retains the ID of the old tracking instance.  Due to the performance limitations of Grounding DINO 1.5, newly detected object instances may already have appeared in preceding frames yet were missed.
Thus, SAM2 conducts a reverse four-second instance tracking for each new object in key frames, thereby achieving full lifecycle instance tracking. In total, we obtain 1.14 million video instance masks from all the egocentric videos.


\subsubsection{Object QA Generation}

In this work, we generate three types of object-related tasks: object captioning, object comprehension QA, and referring video object segmentation. For each instance, we first divide all frames containing the instance into eight equal parts in chronological order. Within each frame group, an instance key frame is selected based on two factors: the size of the instance in the frame and the distance between the instance center and the frame center. Consequently, each instance is associated with eight instance key frames, featuring good instance visibility and diverse viewing angles. Half of these frames have the instance cropped out using a mask, while the other four highlight the instance using a red bounding box and background dimming technique. The final set of object cue images is displayed within the blue box in Fig.~\ref{fig:data_pipeline}.


Due to the limitation of SAM2 in consistent object tracking in egocentric videos, the same instance may be assigned multiple IDs if the instance appears intermittently in the video. We employ an object category filtering method that limits each video to a maximum of two instances per object category, thereby minimizing duplicate instances. The presence of multiple video segments per house leads to repeated occurrences of certain salient objects, causing a pronounced long-tail distribution.
We downsample object categories that occur frequently to prevent extreme object distribution. After the aforementioned filtering, the cue image sets of retained instances are input into Qwen2.5-VL~\cite{qwen25vl}, generating object caption and object comprehension QA through various prompts. It is noteworthy that in the object comprehension QA, counting QA task is particularly unique and requires specially designed prompts. Subsequently, based on each instance's caption and QAs, Qwen3~\cite{qwen3} generates two types of referring expressions: simple referring expressions and situational referring expressions. Simple referring expressions identify objects through a combination of features such as spatial location and category. Situational referring expressions establish a task scenario, requiring the model to infer the instance needed by the user within this context. Each type of QA undergoes manual filtering post-output to ensure data quality. Detailed prompts are provided in the Appendix~\ref{sec:appendix_object_qa}.

\subsubsection{Spatial QA Generation}
\label{sec:spatial_qa}


Unlike object QA, spatial QA requires more precise 3D information concerning the global scene context. Therefore, we utilize MASt3R-SLAM~\cite{mast3r-slam} to reconstruct 3D point clouds from RGB videos and obtain camera extrinsic parameters. Subsequently, by projecting 2D pixel points to 3D coordinates, the segmentation of each instance in the video can be mapped onto the point cloud. However, it is important to note that the world coordinate system established by MASt3R-SLAM for the 3D point cloud is not aligned with the floor. Therefore, the Random Sample Consensus (RANSAC)~\cite{RANSAC} algorithm is implemented to identify inlier points for plane fitting through ten iterative executions. In each iteration, the detected planar surface and its inliers are removed from the point cloud for subsequent plane detection. Given that the initial camera pose was approximately horizontal but not perpendicular to the ground, the ground plane is selected based on minimal angular deviation between its normal vector and the initial camera Y-axis orientation. The point cloud is then aligned to ensure orthogonality between the world coordinate Z-axis and the detected ground plane.


RynnEC dataset encompasses 10 fundamental spatial abilities, each of which is further divided into quantitative and qualitative variants.
We construct spatial QA in a template-based manner. Diverse QA templates are designed according to the characteristics of each task, and the missing attributes within the templates (e.g., distance, height) can be calculated from the 3D point cloud. We denote each instance in the format \texttt{<Object X>}. Furthermore, to obtain purely textual spatial QA pairs, we replace \texttt{<Object X>} with simple referring expressions generated in the above object QA pipeline. These texts are then further refined and diversified using GPT-4o, resulting in the final natural language spatial QA data. With training on these data, RynnEC is able to answer spatial questions in various input forms. Examples of the generated spatial QAs are illustrated in Fig.~\ref{fig:data_pipeline}, and more examples as well as detailed templates are provided in the Appendix~\ref{sec:appendix_spatial_qa}.


Building on insights from prior works~\cite{Spatial-MLLM, SpaceR}, we recognize that spatial cognition tasks are highly challenging. Therefore, in addition to constructing a large-scale video-based spatial QA dataset, we also develop a relatively simpler image-based spatial QA dataset. This combination of tasks with varying levels of difficulty is intended to improve learning efficiency and enhance model robustness. Specifically, we collect 500k indoor images from 39k houses. Leveraging the single-image-to-3D reconstruction and calibration methods from SpatialRGPT~\cite{spatialrgpt}, we obtain the 3D spatial relationships between objects in each image. We then select tasks from the video-based spatial cognition set that can also be addressed via single images, and design corresponding QA templates. The format of the image-based spatial QA is kept consistent with that of the video-based spatial QA.


\subsection{RynnEC-Bench}
\label{sec:RynnEC-Bench}

\begin{figure*}[t]
\includegraphics[width=\textwidth]{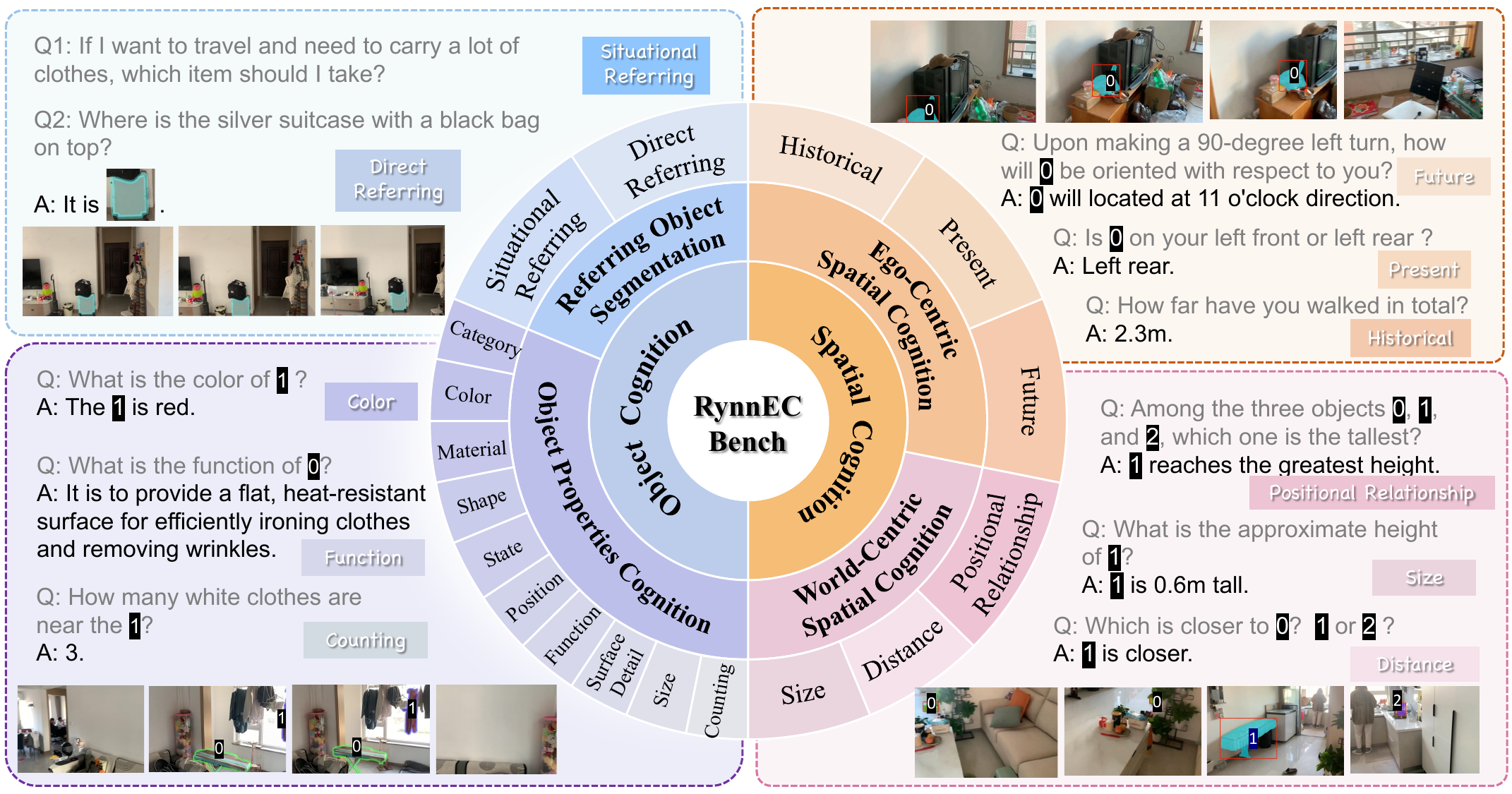}
\caption{\textbf{Overview of embodied cognition dimensions in RynnEC-Bench.} RynnEC-Bench includes two subsets: object cognition and spatial cognition, evaluating a total of 22 embodied cognitive abilities.} 
\label{fig:benchmark}
\end{figure*}




As this work is the first to propose a comprehensive set of fine-grained embodied video tasks, a robust evaluation framework for assessing MLLMs’ overall capabilities in this domain is currently lacking. To address this, we propose RynnEC-Bench, which evaluates fine-grained embodied understanding models from the perspectives of object cognition and spatial cognition in open-world scenarios. Fig.~\ref{fig:benchmark} provides a detailed illustration of the capability taxonomy in RynnEC-Bench.


\subsubsection{Capability Taxonomy}

Object cognition is divided into two tasks: object properties cognition and referring object segmentation. During embodied task execution, robots often require a clear understanding of key objects' functions, locations, quantities, surface details, relationships with the surrounding environment, etc. Accordingly, the object properties recognition tasks comprehensively and meticulously construct questions in these aspects. In the processes of robotic manipulation and navigation, identifying operation instances and target instances is an essential step. Precise instance segmentation in videos serves as the best approach to indicate the positions of these key objects. Specifically, the referring object segmentation task is categorized into direct referring problems and situational referring problems. Direct referring problems involve only combinations of descriptions for the instance, while situational referring problems are set within a scenario, requiring MLLMs to perform reasoning in order to identify the target object.

Spatial cognition requires MLLMs to derive a 3D spatial awareness from egocentric video.
We categorize it into ego-centric and world-centric spatial cognition.
Ego-centric spatial cognition maintains awareness of agent-environment spatial relations and supports spatial reasoning and mental simulation; by temporal scope, we consider past, present, and future cases.
World-centric spatial cognition focuses on understanding the 3D layout and scale of the physical world, which we further evaluate in terms of size, distance, and positional relations.


\subsubsection{Data Balance}
\label{sec:data_balance}
The videos in RynnEC-Bench are collected from ten houses that do not overlap with those in the training set.
When evaluating object cognition, we observe substantial variation in object-category distributions across houses, making results highly sensitive to which houses are sampled.
To mitigate this bias and better reflect real-world deployment, we introduce a physical-world-based evaluation protocol.
We first define a taxonomy of 12 coarse and 119 fine-grained indoor object categories. 
Using GPT-4o, we then estimate an empirical category-frequency distribution by parsing 500,000 indoor images from 39,000 houses; given the scale, this serves as a close approximation to real-world indoor object frequencies.
Finally, we perform frequency-proportional sampling so that the object-category distribution in RynnEC-Bench closely matches the empirical distribution, enabling a more objective and realistic evaluation. 
Specifically, counting questions with answers of 1 or 2 are reduced by 50\% to achieve a more balanced difficulty distribution.
All QA pairs in RynnEC-Bench are further subjected to meticulous human screening to ensure high quality.
Additional implementation details are available in Appendix~\ref{sec:rynnec-bench-appe}.


\subsubsection{Evaluation Framework}

The questions are categorized into three types based on the nature of their answers: numerical questions, textual questions, and segmentation questions. 
For numerical questions such as distance estimation and direction estimation, we directly use the formula to calculate the precise indicators.
For scale-related questions, Mean Relative Accuracy (MRA)~\cite{vsibench, pascal_voc} is used to calculate the scores.
Specifically, given a model's prediction $\hat{y}$, ground truth $y$, and a confidence threshold $\theta$, relative accuracy is calculated by considering $\hat{y}$ correct if the relative error rate, defined as $|\hat{y} - y|/y$, is less than $1 - \theta$. As single-confidence-threshold accuracy only considers relative error within a narrow scope, the MRA averages the relative accuracy across a range of confidence thresholds $\mathcal{C} = \{0.5,\, 0.55,\, \ldots,\, 0.95\}$:
\begin{equation}
MRA = \frac{1}{|\mathcal{C}|} \sum_{\theta \in \mathcal{C}} \mathbb{I} \Bigg( \frac{|\hat{y} - y|}{y} < 1 - \theta \Bigg)
\end{equation}
where $\mathbb{I}(\cdot)$ is the indicator function. 
For angle-related questions, MRA is not suitable due to the cyclic nature of angular measurements. We therefore designed a rotational accuracy metric (RoA). 
\begin{equation}
 RoA = 1 - min \Bigg(\frac{min(|\widehat{y} - y|, 360- |\widehat{y} - y|)}{90} , 1 \Bigg)
\end{equation}
RoA assigns a score only when the angular difference is less than 90 degrees, ensuring consistency in task difficulty across different settings.

Textual questions are further categorized into close-ended and open-ended questions.
For the close-ended part, we prompt GPT-4o to assign a
straightforward binary score of either 0 or 1. For the open-ended part, answers are scored by GPT-4o on a scale from 0 to 1 in increments of 0.2. This question-type-adaptive evaluation approach enables the metrics of RynnEC-Bench to be both precise and consistent.


For segmentation evaluation, prior work~\cite{sa2va, visa} typically reports the $\mathcal{J}\&\mathcal{F}$ measure, combining region-overlap ($\mathcal{J}$) and boundary-accuracy ($\mathcal{F}$) scores. However, the conventional frame-averaged $\mathcal{J}\&\mathcal{F}$ treats empty frames (i.e., frames with no ground-truth mask) in a binary manner: if any predicted mask appears, the frame score is set to 0; otherwise it is set to 1.
This evaluation method fails to account for the actual size of erroneous masks in empty frames, which can have a significant impact on embodied segmentation tasks.
To address this, we propose the Global IoU metric, defined as
\begin{equation}
    \overline{\mathcal{J}} = \frac{\sum_{i=1}^{N} |\mathcal{S}_i \cap \mathcal{G}_i|}{\sum_{i=1}^{N} |\mathcal{S}_i \cup \mathcal{G}_i|},
\end{equation}
where $N$ is the total number of video frames, $\mathcal{S}_i$ denotes the predicted segmentation mask for frame $i$, and $\mathcal{G}_i$ denotes the ground truth mask for frame $i$.
For the boundary accuracy metric $\overline{\mathcal{F}}$, we compute the average only over non-empty frames. The mean of $\overline{\mathcal{J}}$ and $\overline{\mathcal{F}}$, denoted as $\overline{\mathcal{J}}\&\overline{\mathcal{F}}$, provides an accurate reflection of segmentation quality, especially in egocentric videos where the target object appears in relatively few frames.

\begin{figure*}[t]
\includegraphics[width=\textwidth]{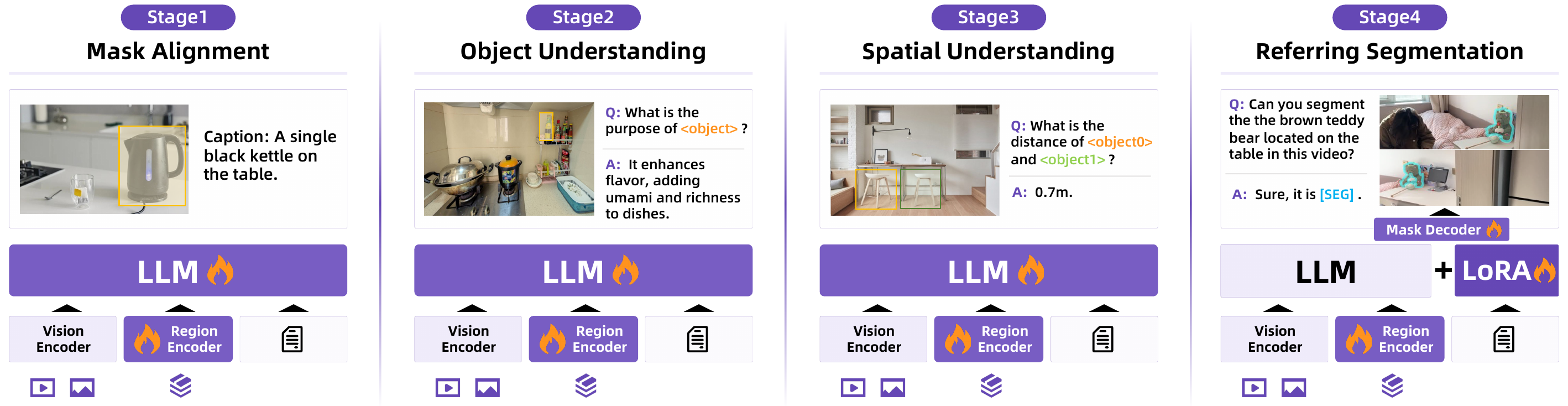}
\caption{
\textbf{Training paradigm of RynnEC.} The model is trained in four progressive stages:
 1) Mask Alignment, 2) Object Understanding, 3) Spatial Understanding, and 4) Referring Segmentation. }
\label{fig:training_stage}
\end{figure*}

\subsection{RynnEC Architecture}
\label{sec:archi}
RynnEC consists of three core components: the foundational vision-language model for basic multimodal comprehension, a region-aware encoder for fine-grained object-centric representation learning, an adaptive mask decoder for video segmentation tasks. Notably, the latter two modules are designed as plug-and-play components with independent parameter spaces, ensuring architectural flexibility and modular extensibility.

\textbf{Foundational Vision-Language Model.} We ultilize VideoLLaMA3-Image~\cite{videollama3} as the foundational vision-language model for RynnEC, which consists of three main modules: a Vision Encoder, the Projector and the Large Language Model (LLM). For the vision encoder, we use VL3-SigLIP-NaViT~\cite{videollama3}, which leverages an any-resolution vision tokenization strategy to flexibly encode images of varying resolutions. As the LLM, we employ Qwen2.5-1.5B-Instruct~\cite{qwen2} and Qwen2.5-7B-Instruct~\cite{qwen2}, enabling scalable trade-offs between performance and computational cost.

\textbf{Region Encoder.}
Egocentric videos often feature cluttered scenes with similar objects that are difficult to distinguish using linguistic cues alone. To address this, we introduce a dedicated object encoder for specific object representation. This facilitates more precise cross-modal alignment during training and enables intuitive, fine-grained user interaction at inference time. Following ~\cite{osprey,videorefer}, we use a simple yet efficient MaskPooling for object tokenization, followed by a two-layer projector to align object features with LLM embedding space. During training, object masks spanning multiple frames in a video are utilized to achieve accurate representations. At inference, the encoder offers flexibility, operating effectively with either single-frame or multi-frame object masks.

\textbf{Mask Decoder.}
Accurate object localization is critical for egocentric video understanding. To incorporate robust visual grounding capabilities without degrading the model’s pretrained performance, we fine-tune the LLM with LoRA. Our mask decoder is based on the architecture of SAM2~\cite{sam2}, which has demonstrated strong generalization capabilities and prior knowledge in purely visual segmentation tasks.
For a given video and the instruction, we adpot a \texttt{[SEG]} token as a specifical token to trigger mask generation for the corresponding visual region. To facilitate this process, an additional linear layer is introduced to align the \texttt{[SEG]} token with SAM2's feature space.

\subsection{Training and Inference}
\label{sec:training}
As illustrated in Fig.~\ref{fig:training_stage}, RynnEC is trained using a progressive four-stage pipeline: 1) Mask Alignment, 2) Object Understanding, 3) Spatial Understanding, and 4) Referring Segmentation. The first three stages are designed to incrementally enhance fine-grained, object-centric understanding, while the final stage focuses on equipping the model with precise object-level segmentation capabilities. This curriculum-based approach ensures gradual integration of visual, spatial, and grounding knowledge without overfitting to a single task. The datasets used in each stage are summarized in Tab.~\ref{tab:datasets_used}. 
The details of each training stage are as follows:

\textbf{1) Mask Alignment.} The goal of this initial stage is to encourage the model to attend to region-specific tokens rather than relying solely on global visual features. We fine-tune both the region encoder and the LLM on a large-scale object-level captioning dataset, where each caption is explicitly aligned with a specific object mask. This alignment training conditions the model to associate object-centric embeddings with corresponding linguistic descriptions, laying the foundation for localized reasoning in later stages.

\textbf{2) Object Understanding.} 
In this stage, the focus shifts to enriching the model’s egocentric object knowledge, encompassing attributes such as color, shape, material, size, and functional properties. The region encoder and the LLM are jointly fine-tuned to integrate this object-level information more effectively into the cross-modal embedding space. This stage is the basic for spatial understanding. 

\textbf{3) Spatial Understanding.} 
Building on the previous stage, this phase equips the model with spatial reasoning abilities, enabling it to understand and reason about the relative positions and configurations of objects within a scene. We use a large amount of spatial QA we generated and the previous stage data as well as general VQA to maintain the ability to follow instructions.

\textbf{4) Referring Segmentation.} In the final stage, we integrate the Mask Decoder module after the LLM to endow the model with fine-grained referring segmentation capabilities. The LLM is fine-tuned via LoRA to minimize interference with its pretrained reasoning abilities. The training data includes not only segmentation-specific datasets but also samples from earlier stages to mitigate catastrophic forgetting. This multi-task mixture ensures that segmentation performance is improved without sacrificing the model’s object and spatial understanding.


\begin{table}[t]
\centering
\renewcommand{\arraystretch}{1.2} 
\footnotesize
\caption{\textbf{Datasets used at four training stages.} \textbf{IM} and \textbf{OM} indicate whether the task involves the input mask and output mask, respectively.}
\label{tab:datasets_used}
\resizebox{\textwidth}{!}{%
\begin{tabularx}{\textwidth}{@{}>{\centering\arraybackslash}m{3cm}|>{\centering\arraybackslash}m{3.5cm}|>{\centering\arraybackslash}m{0.5cm}|>{\centering\arraybackslash}m{0.5cm}|>{\centering\arraybackslash}m{1.0cm}|X@{}}
\toprule
\textbf{Training Stage}  & \textbf{Task} & \textbf{IM} & \textbf{OM} & \textbf{\# Samples} & \textbf{Datasets} \\ 
\midrule

\multirowcell{2}[-2.5ex][c]{\textbf{Mask Alignment}\\\textbf{(Stage-1)}}
    & General Mask Captioning 
        & \textcolor{green}{\Checkmark} 
        & \textcolor{red}{\XSolidBrush} 
        & 1.17M 
        & RefCOCO~\cite{refcoco,refcocog}, VideoRefer-Caption~\cite{videorefer}, DAM~\cite{dam}, Osprey-Caption~\cite{osprey}, MDVP-Data~\cite{lin2024draw}, HC-STVG~\cite{hc_stvg} \\
    & \cellcolor{black!5}Scene Instance Captioning 
        & \cellcolor{black!5}\textcolor{green}{\Checkmark} 
        & \cellcolor{black!5}\textcolor{red}{\XSolidBrush} 
        & \cellcolor{black!5}0.14M 
        & \cellcolor{black!5}RynnEC-Caption \\ 
\midrule

\multirowcell{2}[0.0ex][c]{\textbf{Object Understanding}\\\textbf{(Stage-2)}}
    & Basic Properties QA 
        & \textcolor{green}{\Checkmark} 
        & \textcolor{red}{\XSolidBrush} 
        & 1.49M 
        & RynnEC-Object \\
    & \cellcolor{black!5}Object-Centric Counting 
        & \cellcolor{black!5}\textcolor{green}{\Checkmark} 
        & \cellcolor{black!5}\textcolor{red}{\XSolidBrush} 
        & \cellcolor{black!5}0.25M 
        & \cellcolor{black!5}RynnEC-Counting \\ 
\midrule

\multirowcell{2}[-8ex][c]{\textbf{Spatial Understanding}\\\textbf{(Stage-3)}}
    & Our Stage-2 
        & \textcolor{green}{\Checkmark} 
        & \textcolor{red}{\XSolidBrush} 
        & 0.30M 
        & RynnEC-Object, RynnEC-Counting \\
    & \multirow{2}{*}{\cellcolor{black!5}\centering Spatial QA} 
        & \cellcolor{black!5}\textcolor{green}{\Checkmark} 
        & \cellcolor{black!5}\textcolor{red}{\XSolidBrush} 
        & \cellcolor{black!5}0.60M 
        & \cellcolor{black!5}RynnEC-Spatial (Image), RynnEC-Spatial (Video)\\
    &  \cellcolor{black!5}
        & \cellcolor{black!5}\textcolor{red}{\XSolidBrush} 
        & \cellcolor{black!5}\textcolor{red}{\XSolidBrush} 
        & \cellcolor{black!5}0.54M 
        & \cellcolor{black!5}VLM-3R-Data~\cite{vlm-3r} \\
    & General VQA 
        & \textcolor{red}{\XSolidBrush} 
        & \textcolor{red}{\XSolidBrush} 
        & 0.74M 
        & LLaVA-OV-SI~\cite{llava-ov}, LLaVA-Video~\cite{llava-video}, ShareGPT-4o-video~\cite{sharegpt4video}, VideoGPT-plus~\cite{videogpt+}, FineVideo~\cite{FineVideo}, CinePile~\cite{cinepile}, ActivityNet~\cite{activitynet}, YouCook2~\cite{youcook2}, LLaVA-SFT~\cite{llava} \\ 
\midrule

\multirowcell{2}[-10ex][c]{\textbf{Referring Segmentation}\\\textbf{(Stage-4)}}
    & \cellcolor{black!5}Our Stage-2 \& Stage-3
        & \cellcolor{black!5}\textcolor{green}{\Checkmark} 
        & \cellcolor{black!5}\textcolor{red}{\XSolidBrush} 
        & \cellcolor{black!5}0.60M 
        & \cellcolor{black!5}RynnEC-Object, RynnEC-Counting, RynnEC-Spatial \\
    & General Segmentation 
        & \textcolor{red}{\XSolidBrush} 
        & \textcolor{green}{\Checkmark} 
        & 0.32M 
        & ADE20K~\cite{ade20k}, COCOStuff~\cite{cocostuff}, Mapillary~\cite{mapillary}, PACO-LVIS~\cite{paco}, PASCAL-Part~\cite{pascal-part} \\
    & \cellcolor{black!5}Embodied Segmentation 
        & \cellcolor{black!5}\textcolor{red}{\XSolidBrush} 
        & \cellcolor{black!5}\textcolor{green}{\Checkmark} 
        & \cellcolor{black!5}0.31M 
        & \cellcolor{black!5}RynnEC-Segmentation \\ 
    & General VQA 
        & \textcolor{red}{\XSolidBrush} 
        & \textcolor{red}{\XSolidBrush} 
        & 0.80M 
        & LLaVA-OV-SI~\cite{llava-ov}, LLaVA-Video~\cite{llava-video}, ShareGPT-4o-video~\cite{sharegpt4video}, VideoGPT-plus~\cite{videogpt+}, FineVideo~\cite{FineVideo}, CinePile~\cite{cinepile}, ActivityNet~\cite{activitynet}, YouCook2~\cite{youcook2}, LLaVA-SFT~\cite{llava} \\
\bottomrule
\end{tabularx}}
\end{table}

\clearpage
\section{Experiments}
\noindent







\begin{table}[t]
\caption{\textbf{Main evaluation results on RynnEC-Bench.} We evaluate in two major categories: Object Cognition and Spatial Cognition. \textbf{DR} and \textbf{SR} represent Direct Referring and Situational Referring. \textbf{PR} represents Positional Relationship. }
\label{tab:main_result}
\resizebox{\columnwidth}{!}{%
\begin{tabular}{lcccccccccccc}
\Xhline{2\arrayrulewidth}
\multicolumn{1}{l|}{\multirow{3}{*}{\textbf{Model}}} & \multicolumn{1}{c|}{\multirow{3}{*}{\begin{tabular}[c]{@{}c@{}}\textbf{Overall}\\ \textbf{Mean}\end{tabular}}} & \multicolumn{4}{c|}{\textbf{Object Cognition}} & \multicolumn{7}{c}{\textbf{Spatial Cognition}} \\ \cline{3-13} 
\multicolumn{1}{l|}{} & \multicolumn{1}{c|}{} & \multirow{2}{*}{\begin{tabular}[c]{@{}c@{}}\textbf{Object} \\ \textbf{Properties}\end{tabular}} & \multicolumn{2}{c|}{\textbf{Segmentation}} & \multicolumn{1}{c|}{\multirow{2}{*}{\textbf{Mean}}} & \multicolumn{3}{c}{\textbf{Ego-Centric}} & \multicolumn{3}{c|}{\textbf{World-Centric}} & \multirow{2}{*}{\textbf{Mean}} \\
\multicolumn{1}{l|}{} & \multicolumn{1}{c|}{} &  & \multicolumn{1}{r}{\textbf{DR}} & \multicolumn{1}{c|}{\textbf{SR}} & \multicolumn{1}{c|}{} & \textbf{His.} & \textbf{Pres.} & \textbf{Fut.} & \textbf{Size} & \textbf{Dis.} & \multicolumn{1}{c|}{\textbf{PR}} &  \\\Xhline{2\arrayrulewidth}
\rowcolor{myred!10}
\multicolumn{13}{c}{\textit{Proprietary Generalist MLLMs}} \\ \Xhline{2\arrayrulewidth}
\multicolumn{1}{l|}{GPT-4o~\cite{gpt4o}} & \multicolumn{1}{c|}{28.3} & 41.1 & \textbf{---} &  \multicolumn{1}{c|}{\textbf{---}} & \multicolumn{1}{c|}{33.9} & 13.4 & 22.8 & 6.0 & 24.3 & 16.7 & \multicolumn{1}{c|}{36.1} & 22.2 \\
\multicolumn{1}{l|}{GPT-4.1~\cite{gpt4o}} & \multicolumn{1}{c|}{33.5} & 45.9 & \textbf{---} &  \multicolumn{1}{c|}{\textbf{---}} & \multicolumn{1}{c|}{37.8} & 17.2 & 27.6 & 6.1 & 35.9 & 30.4 & \multicolumn{1}{c|}{45.7} & 28.8 \\
\multicolumn{1}{l|}{Seed1.5-VL~\cite{seed1.5VL}} & \multicolumn{1}{c|}{34.7} & 52.1 & \textbf{---} &  \multicolumn{1}{c|}{\textbf{---}} & \multicolumn{1}{c|}{42.8} & 8.2 & 27.7 & 4.3 & 32.9 & 19.1 & \multicolumn{1}{c|}{27.9} & 26.1 \\
\multicolumn{1}{l|}{Genimi-2.5 Pro~\cite{gemini2—5}} & \multicolumn{1}{c|}{45.5} & \textbf{64.0} & \textbf{---} & \multicolumn{1}{c|}{\textbf{---}} & \multicolumn{1}{c|}{52.7} & 9.3 & 36.7 & 8.1 & 47.0 & 29.9 & \multicolumn{1}{c|}{69.3} & 37.8 \\ \Xhline{2\arrayrulewidth}
\rowcolor{myyellow!30}
\multicolumn{13}{c}{\textit{Open-source Generalist MLLMs}} \\ \Xhline{2\arrayrulewidth}
\multicolumn{1}{l|}{VideoLLaMA3-7B~\cite{videollama3}} & \multicolumn{1}{c|}{27.3} & 36.7 & \textbf{---} & \multicolumn{1}{c|}{\textbf{---}} & \multicolumn{1}{c|}{30.2} & 5.1 & 26.8 & 1.2 & 30.0 & 19.0 & \multicolumn{1}{c|}{34.9} & 24.1 \\
\multicolumn{1}{l|}{InternVL3-78B~\cite{InternVL3}} & \multicolumn{1}{c|}{29.0} & 45.3 & \textbf{---} & \multicolumn{1}{c|}{\textbf{---}} & \multicolumn{1}{c|}{37.3} & 9.0 & 31.8 & 2.2 & 10.9 & 30.9 & \multicolumn{1}{c|}{26.0} & 20.0 \\ 
\multicolumn{1}{l|}{Qwen2.5-VL-72B~\cite{qwen25vl}} & \multicolumn{1}{c|}{36.4} & 54.2 & \textbf{---} & \multicolumn{1}{c|}{\textbf{---}} & \multicolumn{1}{c|}{44.7} & 11.3 & 24.8 & 7.2 & 27.2 & 22.9 & \multicolumn{1}{c|}{83.7} & 27.4 \\
\Xhline{2\arrayrulewidth}
\rowcolor{mygreen!50}
\multicolumn{13}{c}{\textit{Open-source Object-Level MLLMs}} \\ \Xhline{2\arrayrulewidth}
\multicolumn{1}{l|}{DAM-3B~\cite{dam}} & \multicolumn{1}{c|}{15.6} & 22.2 & \textbf{---} & \multicolumn{1}{c|}{\textbf{---}} & \multicolumn{1}{c|}{18.3} & 2.8 & 14.1 & 1.3 & 28.7 & 6.1 & \multicolumn{1}{c|}{18.3} & 12.6 \\
\multicolumn{1}{l|}{VideoRefer-VL3-7B~\cite{videorefer}} & \multicolumn{1}{c|}{32.9} & 44.1 & \textbf{---} & \multicolumn{1}{c|}{\textbf{---}} & \multicolumn{1}{c|}{36.3} & 5.8 & 29.0 & 6.1 & 38.1 & 30.7 & \multicolumn{1}{c|}{28.8} & 29.3 \\ 
\Xhline{2\arrayrulewidth}
\rowcolor{myblue}
\multicolumn{13}{c}{\textit{Referring Video Object Segmentation MLLMs}} \\ \Xhline{2\arrayrulewidth}
\multicolumn{1}{l|}{Sa2VA-4B~\cite{sa2va}} & \multicolumn{1}{c|}{4.9} & 5.9 & 35.3 & \multicolumn{1}{c|}{14.8} & \multicolumn{1}{c|}{9.4} & 0.0 & 0.0 & 1.3 & 0.0 & 0.0 & \multicolumn{1}{c|}{0.0} & 0.0 \\ 
\multicolumn{1}{l|}{VideoGlaMM-4B~\cite{videoglamm}} & \multicolumn{1}{c|}{9.0} & 16.4 & 5.8 & \multicolumn{1}{c|}{4.2} & \multicolumn{1}{c|}{14.4} & 4.1 & 4.7 & 1.4 & 0.8 & 0.0 & \multicolumn{1}{c|}{0.3} & 3.2 \\ 
\multicolumn{1}{l|}{RGA3-7B~\cite{RGA3}} & \multicolumn{1}{c|}{10.5} & 15.2 & 32.8 & \multicolumn{1}{c|}{23.4} & \multicolumn{1}{c|}{17.5} & 0.0 & 5.5 & 6.1 & 1.2 & 0.9 & \multicolumn{1}{c|}{0.0} & 3.0 \\
\Xhline{2\arrayrulewidth}
\rowcolor{myorange}
\multicolumn{13}{c}{\textit{Open-source Embodied MLLMs}} \\ \Xhline{2\arrayrulewidth}
\multicolumn{1}{l|}{RoboBrain-2.0-32B~\cite{robobrain2.0}} & \multicolumn{1}{c|}{24.2} & 25.1 & \textbf{---} & \multicolumn{1}{c|}{\textbf{---}} & \multicolumn{1}{c|}{20.7} & 8.8 & 34.1 & 0.2 & 37.2 & 30.4 & \multicolumn{1}{c|}{3.6} & 28.0 \\ 
\Xhline{2\arrayrulewidth}
\rowcolor{mygray!20}
\multicolumn{1}{l|}{\textbf{RynnEC-2B}} & \multicolumn{1}{c|}{\underline{54.4}} & 59.3 &  \textbf{46.2} & \multicolumn{1}{c|}{\textbf{36.9}} & \multicolumn{1}{c|}{\underline{56.3}} & \underline{30.1} & \underline{47.2} & \textbf{23.8} & \textbf{67.4} & \underline{31.2} & \multicolumn{1}{c|}{\underline{85.8}} & \underline{52.3} \\ 
\rowcolor{mygray!20}
\multicolumn{1}{l|}{\textbf{RynnEC-7B}} & \multicolumn{1}{c|}{\textbf{56.2}} & \underline{61.4} &  \underline{45.3} & \multicolumn{1}{c|}{\underline{36.1}} & \multicolumn{1}{c|}{\textbf{57.8}} & \textbf{40.9} & \textbf{50.2} & \underline{22.3} & \underline{67.1} & \textbf{39.2} & \multicolumn{1}{c|}{\textbf{89.7}} & \textbf{54.5} \\ \Xhline{2\arrayrulewidth}
\end{tabular}%
}
\end{table}

\subsection{Implementation Details}
\subsubsection{Training}
In this part, we briefly introduce the implementation details of each training stage. For all stages, we adopt the cosine learning rate scheduler. The warm up ratio of the learning rate is set as 0.03. 
The maximum token length is set to 16384, while the maximum token length for vision tokens is set to 8192. In Stage 1, both the vision encoder and the LLM are initialized with pretrained weights from VideoLLaMA3-Image. During this stage, we train the LLM, the projector, and the region encoder, using learning rates of $1 \times 10^{-5}$, $1 \times 10^{-5}$, and $4 \times 10^{-5}$, respectively. In Stages 2 and 3, the learning rates for the LLM, projector, and region encoder are adjusted to $4 \times 10^{-5}$, $1 \times 10^{-5}$, and $1 \times 10^{-5}$, respectively. In the final stage, the LLM is fine-tuned using LoRA with the same learning rates as in Stage 3. The learning rate of Mask Decoder is set to $4 \times 10^{-5}$.

\subsubsection{Evaluation}

We present a comprehensive evaluation of five MLLM categories on RynnEC-Bench, including both general-purpose models and those fine-tuned for region-level understanding and segmentation. For models that do not accept direct region-based inputs, we uniformly highlight target objects using bounding boxes in the video. Multiple objects are distinguished by different colored boxes, which are referenced in the question prompt.
We observe that general-purpose MLLMs are incapable of localizing objects in videos; thus, only specialist models fine-tuned for this ability are evaluated on the RynnEC-Bench segmentation subset. To ensure a consistent evaluation protocol, videos are sampled at 1 fps up to a maximum of 30 frames. If the initial sampling exceeds the 30-frame limit, these target-containing frames are kept, and the remaining frames are selected via uniform sampling from the rest of the video.

\begin{figure*}[t]
\includegraphics[width=\textwidth]{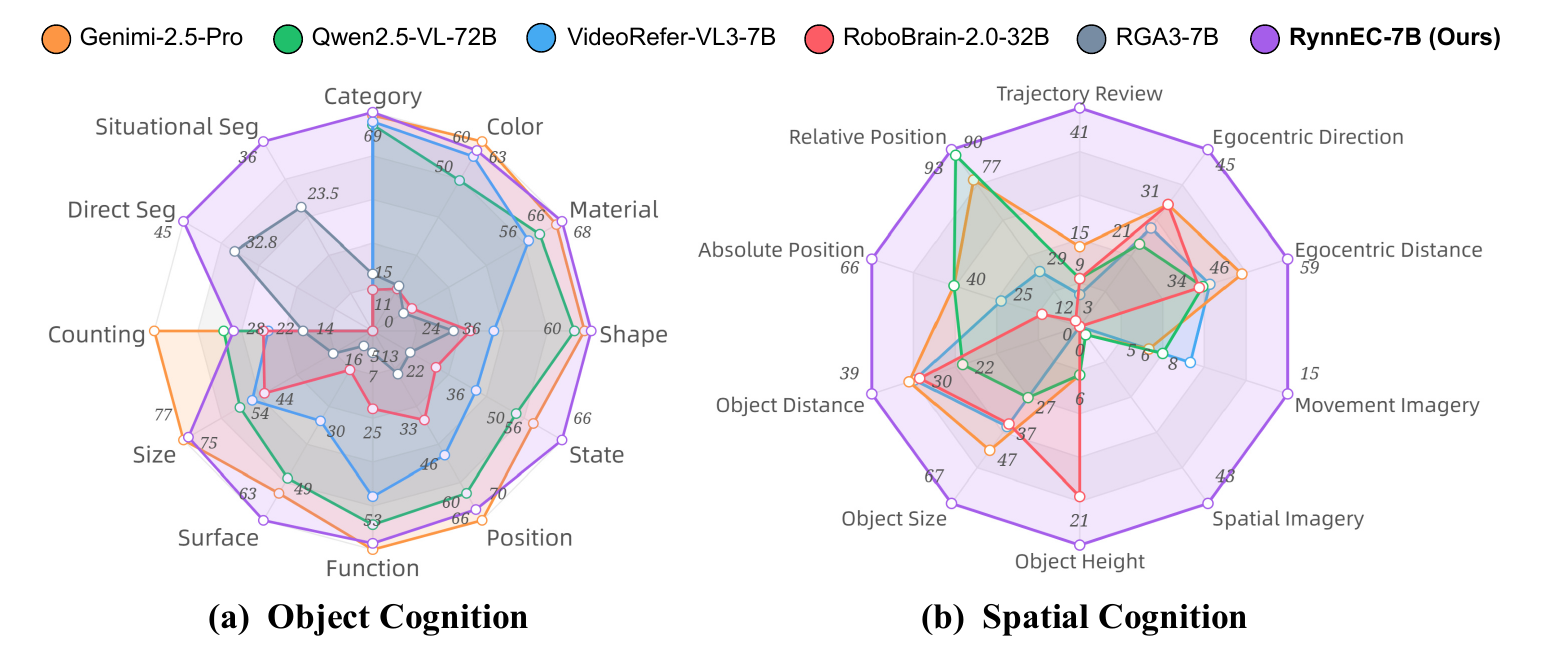}
\caption{\textbf{More granular assessments of object cognition and spatial cognition.} We compare the best-performing MLLM from each category with our RynnEC-7B.} 
\label{fig:lader}
\end{figure*}

\subsection{Embodied Cognition Evaluation}
\subsubsection{Main Results} \label{main-result}

Tab.~\ref{tab:main_result} presents the evaluation results of our RynnEC model and five categories of related MLLMs on the RynnEC-Bench. Although the RynnEC model contains only 7B parameters, it demonstrates robust embodied cognitive abilities, outperforming even the most advanced proprietary model, Gemini-2.5 Pro~\cite{gemini2—5}, by 10.7 points. Moreover, RynnEC achieves both balanced and superior performance across various tasks. For object cognition, RynnEC achieved a score of 61.4 and possesses the ability to both understand and segment objects. In terms of spatial cognition, RynnEC achieves a score of 54.5, which is 44.2\% higher than that of Gemini-2.5 Pro. To support resource-constrained settings, we present a 2B-parameter RynnEC that delivers markedly lower inference latency while maintaining near-parity performance ($< 2$ percentage points drop), enabling on-device deployment for embodied applications. In the following sections, we will introduce the performance of different types of MLLMs on RynnEC-Bench in detail.

\paragraph{Proprietary Generalist MLLMs}
Among the four leading proprietary generalist MLLMs evaluated, Gemini-2.5 Pro establishes a clear lead with an overall score of 45.5. This represents a substantial performance margin of 25\% over the best open-source generalist MLLM and 38.3\% over the premier open-source object-level MLLM. Even more notably, it achieves a remarkable score of 37.8 in the notoriously difficult domain of spatial cognition. This finding provides compelling evidence that spatial awareness can emerge as a byproduct of extensive training on video comprehension tasks.

\paragraph{Open-source Generalist MLLMs}
Qwen2.5-VL-72B~\cite{qwen25vl} exhibits outstanding performance, achieving a score of 36.4 and surpassing GPT-4.1~\cite{gpt4o}. This suggests that, in specialized capabilities such as embodied cognition, the gap between open-source and proprietary MLLMs has been significantly narrowed. Furthermore, we observe that Qwen2.5-VL and InternVL3~\cite{InternVL3} demonstrate superior performance in positional relationship (PR) and distance perception tasks, respectively, even outperforming Gemini-2.5 Pro. Such pronounced differences in various aspects of spatial cognition may be attributed to the distribution of training data.

\paragraph{Open-source Object-Level MLLMs}
These MLLMs are capable of accepting region masks as input, enabling more direct localization of target objects and facilitating finer-grained object perception. VideoRefer-VL3-7B~\cite{videorefer} is a model fine-tuned from the base model VideoLLaMA3-7B~\cite{videollama3}. As shown in Tab.~\ref{tab:main_result}, VideoRefer-VL3-7B consistently outperforms VideoLLaMA3-7B in both object cognition and spatial cognition tasks. This demonstrates that, in embodied scenarios, integrating mask understanding within the model is superior to explicit visual prompting.

\paragraph{Referring Video Object Segmentation MLLMs}
Recently, several studies have applied MLLMs to object segmentation tasks while retaining the original multimodal understanding capabilities of MLLMs. However, the best-performing model, RGA3-7B~\cite{RGA3}, achieves only 15.2 points on the object properties task. Although these MLLMs can still address some general video understanding tasks, their task generalization ability is significantly diminished following segmentation training. In contrast, our RynnEC model, which is specifically designed for embodied scenarios, maintains strong object and spatial understanding capabilities even after segmentation training.

\paragraph{Open-source Embodied MLLMs}
With the growing demand for highly generalizable cognitive abilities in the field of embodied intelligence, a number of studies have begun to develop MLLMs specifically tailored for embodied scenarios. A representative model is RoboBrain-2.0~\cite{robobrain2.0}, which achieves 24.2 even worse than general-purpose video models such as VideoLLaMA3-7B. There are two primary reasons for this: (1) Loss of object cognition: Embodied MLLMs typically emphasize spatial perception and task planning abilities, but tend to overlook the importance of detailed object understanding. (2) Lack of fine-grained perceptual capability: In egocentric videos, RoboBrain-2.0 demonstrates limited ability to interpret region-level features.

\subsubsection{Object Cognition}
Fig.~\ref{fig:lader} (a) presents a more comprehensive evaluation of RynnEC’s capability in object properties cognition from multiple dimensions. Since most object properties cognition abilities are encompassed by general video understanding skills, Gemini-2.5-Pro exhibits superior performance across various competencies. However, due to the high edge deployment requirements of embodied MLLMs, the inference speed of these large-scale models becomes a bottleneck. With only 7B parameters, RynnEC achieves object properties cognition comparable to that of Gemini-2.5-Pro in most categories. Notably, for attributes such as surface details, object state, and object shape, RynnEC-2B even surpasses all other MLLMs. Moreover, most MLLMs lack video object segmentation capabilities, whereas dedicated segmentation MLLMs often sacrifice understanding abilities. RynnEC, while maintaining strong comprehension capabilities, achieves 30.9\% and 57.7\% improvements over state-of-the-art segmentation MLLMs in direct referring and situational referring object segmentation tasks, respectively.

\subsubsection{Spatial Cognition}

Fig.~\ref{fig:lader} (b) demonstrates RynnEC’s spatial cognition capabilities through more fine-grained tasks. As spatial abilities have not been formally defined or systematically explored in previous work, different MLLMs only exhibit strengths in a limited set of specific skills. Overall, spatial cognition abilities such as Spatial Imagery, Movement Imagery, and Trajectory Review are typically absent in prior MLLMs. In contrast, RynnEC possesses a more comprehensive set of spatial abilities, which can facilitate embodied agents in developing spatial awareness within complex environments.



\begin{figure*}[t]
\centering
\begin{minipage}[t]{0.6\textwidth}
\vspace{0pt}
\centering
\includegraphics[width=\linewidth]{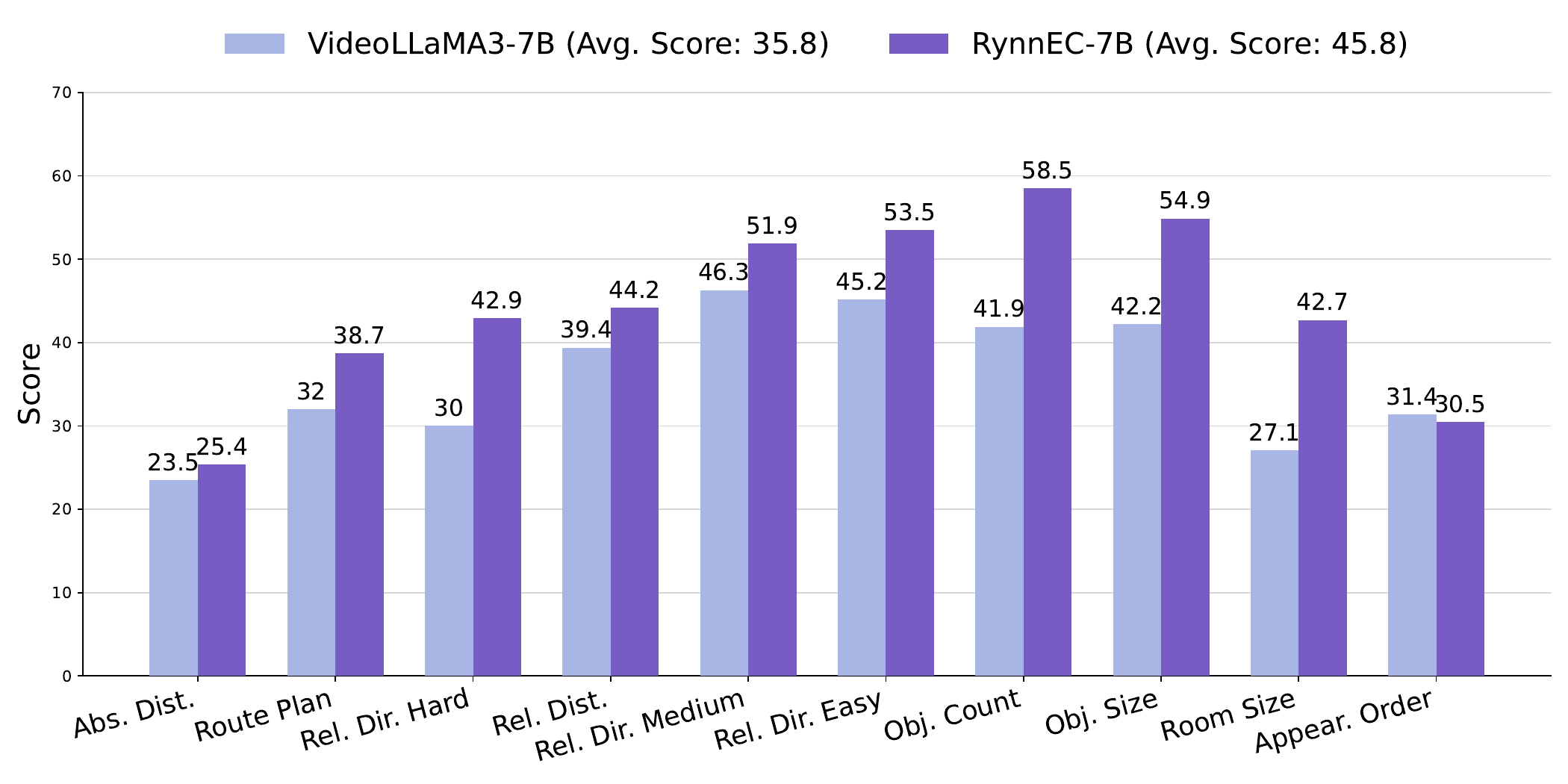}
\end{minipage}\hfill
\begin{minipage}[t]{0.37\textwidth}
\vspace{0pt}
\centering
\begin{tabular}{l|c}
\Xhline{2\arrayrulewidth}
\textbf{Models} & \textbf{VSI-Bench} \\ \Xhline{2\arrayrulewidth}
Qwen2.5-VL-7B~\cite{qwen25vl} & 35.9 \\
InternVL3-8B~\cite{InternVL3} & 42.1 \\
GPT-4o~\cite{gpt4o} & 43.6 \\
Magma-8B~\cite{magma} & 12.7 \\
Cosmos-Reason1-7B~\cite{cosmos} & 25.6 \\
VeBrain-8B~\cite{VeBrain} & 26.3 \\
RoboBrain-7B-1.0~\cite{robobrain1.0} & 31.1 \\
RoboBrain-7B-2.0~\cite{robobrain2.0} & 36.1 \\
M2-Reasoning-7B~\cite{M2-reasoning} & 42.3 \\ 
ViLaSR~\cite{wu2025reinforcing} & 45.4 \\
\Xhline{2\arrayrulewidth}
\rowcolor{mygray!20}
\textbf{RynnEC-7B} & \textbf{45.8} \\ \Xhline{2\arrayrulewidth}
\end{tabular}
\end{minipage}
\caption{\textbf{Performance on VSI-Bench~\cite{vsibench}.} Left: per-subtask comparison with VideoLLaMA3, the base model of our RynnEC. Right: overall comparison with generalist MLLMs and embodied MLLMs without explicit 3D encoding.}
\label{fig:vsi_scalability}
\end{figure*}

\subsection{Generalization and Scalability}



To investigate the generalizability of RynnEC, we conduct experiments on VSI-Bench~\cite{vsibench}, a purely textual spatial intelligence benchmark. As shown in Fig.~\ref{fig:vsi_scalability}, RynnEC-7B consistently surpasses VideoLLaMA3-7B across almost all capability dimensions. Notably, RynnEC is trained with a mask-centric spatial awareness paradigm, whereas all tasks in VSI-Bench involve purely textual spatial reasoning. This demonstrates that spatial awareness need not be constrained by the modality of representation, and spatial reasoning abilities can be effectively transferred across modalities. 
Further observation reveals substantial performance gains of RynnEC on the Route Planning task, despite this task not being included during training. This indicates that the navigation performance of embodied agents is currently constrained by foundational spatial perception capabilities, such as the understanding of direction, distance, and spatial relationships. Only with robust foundational spatial cognition can large embodied models achieve superior performance in high-level planning and decision-making tasks. Compared to other embodied MLLMs of comparable size, RynnEC-7B also achieves a leading score of 45.8.

Certain tasks, such as object segmentation and movement imagery, remain significant challenges for RynnEC. We hypothesize that the suboptimal performance on these tasks stems primarily from insufficient training data. To validate this, we conduct an empirical analysis of data scalability across different task categories. As the data volume increases progressively from 20\% to 100\%, the model's performance on all tasks improves steadily. This observation motivates further expansion of the dataset to enhance RynnEC’s spatial reasoning capabilities. However, it is noteworthy that the marginal gains diminish as data volume grows, indicating a decreasing return on scale. Investigating strategies to enhance data diversity in order to sustain this scaling behavior remains a critical open challenge for future research.


\begin{figure*}[t]
\includegraphics[width=\textwidth]{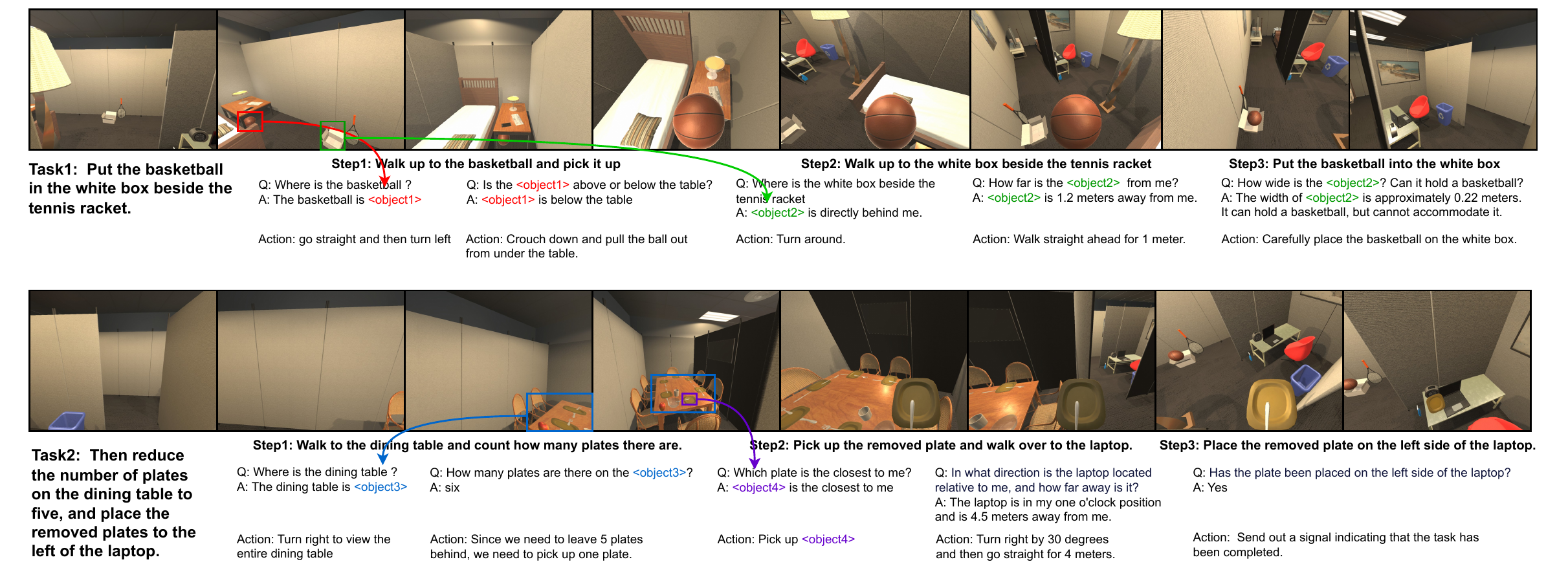}
\caption{
\textbf{The example of RynnEC assisting robots in performing long-range tasks.} The robot accomplishes the two designated tasks within the RoboTHOR simulator~\cite{robothor}. RynnEC facilitates the robot in achieving fine-grained environmental cognition throughout the task execution.} 
\label{fig:task_experiment}
\end{figure*}

\subsection{Embodied Application}

Recently, some works~\cite{efficienteqa, aic} have leveraged MLLMs as the "brain" to assist robots in planning tasks, perceiving environments, and making decisions. However, current MLLMs lack key capabilities such as spatial awareness, fine-grained perception, and instance localization, which restricts these applications to limited and simple tasks. As illustrated in Fig.~\ref{fig:task_experiment}, RynnEC demonstrates the potential to assist robots in accomplishing long-horizon tasks within complex environments. From two real-time tasks performed by the robot equipped with RynnEC, we observe the following roles that RynnEC plays in task execution: (1) Fine-grained object localization and understanding enable robots to more quickly identify target objects and assess their states; (2) Direction and distance perception of targets improves navigation efficiency and precision; (3) Spatial scale estimation empowers robots to perform more delicate manipulations; (4) Counting ability facilitates the completion of tasks requiring mathematical reasoning. It is important to emphasize that the role of RynnEC in embodied tasks is far from limited to these examples. We hope that more researchers will integrate RynnEC models into robotic systems across a wide range of tasks, thereby advancing embodied intelligence toward more valuable real-world applications.


\section{Conclusion and Future Works}
\noindent




In this paper, we introduce RynnEC, a Video MLLM for embodied cognition. 
Through the architectural design of a region encoder and mask decoder, RynnEC achieves flexible, fine-grained visual interaction.
Meanwhile, RynnEC demonstrates robust object and spatial cognitive abilities with compact size.  To address the limitations of available scene data, we employ a data generation pipeline that relies solely on RGB videos. Furthermore, to supplement the lack of fine-grained embodied cognition benchmarks, we propose RynnEC-Bench, which covers 22 categories of object and spatial cognitive abilities. During training, RynnEC progressively integrates diverse skills through a four-stage capability injection process. Importantly, we advocate that fine-grained video-based visual understanding is key to achieving generalizable cognition in the physical world. RynnEC will enable robots to accomplish more precise cognitive tasks, thereby advancing the practical development of embodied intelligence.

We regard RynnEC as a foundational step toward developing a general embodied intelligence model. Looking ahead, we plan to further advance RynnEC along two primary directions.

\begin{itemize}[leftmargin=*,topsep=0pt]
    \item \textbf{Enhancing Reasoning Capabilities:} Robust visual reasoning is essential for solving any complex embodied task. An important research direction is how to effectively integrate RynnEC’s diverse abilities to perform joint reasoning, thereby enabling the resolution of higher-level embodied problems.
    \item \textbf{Unified Perception and Planning Framework:} Recent studies~\cite{robobrain2.0} have started to explore training unified embodied intelligence models that combine perception and planning. However, these approaches are limited in their ability to facilitate fine-grained, video-based visual interactions. In the future, we aim to endow RynnEC with more flexible planning abilities and integrate it with VLA models to form a closed-loop embodied system.
\end{itemize}

\bibliographystyle{assets/plainnat}
\bibliography{paper}

\newpage
\beginappendix
\appendix
\renewcommand{\thesection}{\Alph{section}}
\renewcommand{\thesubsection}{\Alph{subsection}}

\subsection{Implementation Details for Data Pipeline}
\subsubsection{Instance Segmentation}

As described in Section~\ref{sec:instance_seg}, instance segmentation and tracking in videos require a three-stage collaborative process. The first stage involves the extraction of an object list, which should comprehensively include the names of all objects present in the video scene. After evaluating multiple approaches, we find that directly leveraging Qwen2.5-VL to extract object category names from video frames achieves the highest efficiency and accuracy. Specifically, we uniformly sample 16 frames from each video, dividing them into two groups: even-numbered frames and odd-numbered frames. Each group is then processed independently by Qwen2.5-VL to generate a list of object category names. The prompt used to guide the extraction of the object list is in Tab.~\ref{tab:ram_prompt}.

\lstset{
    framesep = 20pt,
    rulesep = 10pt,
    backgroundcolor = \color[RGB]{245,245,244},
    breaklines = true,
    breakindent = 0pt,
    basicstyle = \ttfamily\small,
    escapeinside = {(*@}{@*)} 
}

\begin{table}[h]
\centering
\begin{tcolorbox}[colback=gray!10,
                  colframe=black,
                  width=16.5cm,
                  arc=3mm, auto outer arc,
                 ]
\textbf{\color{red}System Prompt:} Please analyze the image sequence captured as I move through an indoor environment and provide a concise list of major distinct physical objects that can be detected and segmented in these scenes. You need to pay attention to the following points 
(1) Focus on tangible items such as furniture, appliances, and tools. Avoid nouns that denote locations and rooms like "kitchen" or "bedroom". 
(2) Limit the list to a maximum of 20 objects, and avoid including specific components or parts of these objects. 
(3) Ensure the list does not have duplicates.     
Your output must be a series of nouns separated by semicolons
\end{tcolorbox}
\caption{Prompts for object list extraction.}
\label{tab:ram_prompt}
\end{table}


During our experimentation, we observe that Qwen2.5-VL occasionally produces repeated instances of the same object name or phrases sharing the same object name as a prefix. To address this, we apply a post-processing step to remove duplicate and semantically similar phrases from the model outputs, thereby ensuring the diversity and conciseness of the object list. The final object list is obtained by taking the union of the results from the odd-numbered and even-numbered frame groups, yielding a more comprehensive and robust set of detected objects. Furthermore, generic scene-level categories such as "wall" and "floor" are explicitly excluded from the final object list, as they are not considered relevant instances for downstream instance-level tracking and segmentation tasks.

\subsubsection{Object QA Generation}
\label{sec:appendix_object_qa}

We generate three categories of object-related tasks: object caption, object comprehension QA, and referring video object segmentation. The pre-annotation prompts for object caption and object comprehension QA are presented in Tab.~\ref{tab:caption_prompt}. Both tasks take as input a set of keyframes in which the target object is highlighted; the only difference lies in the task-specific instruction prompts.

The referring video object segmentation task requires generating unique referring expressions for objects. We aggregate the QAs generated in the previous stage for each object, representing various attributes of the object. Subsequently, Qwen3 utilizes these QAs to generate both direct referring expressions and situational referring expressions. The specific prompt is shown in Tab.~\ref{tab:seg_prompt}.

\clearpage

\lstset{
    framesep = 20pt,
    rulesep = 10pt,
    backgroundcolor = \color[RGB]{245,245,244},
    breaklines = true,
    breakindent = 0pt,
    basicstyle = \ttfamily\small,
    escapeinside = {(*@}{@*)} 
}

\begin{table}[H]
\centering
\begin{tcolorbox}[colback=gray!10,
                  colframe=black,
                  width=16.5cm,
                  arc=3mm, auto outer arc,
                 ]
\textbf{\color{red}Crop Image Prompt:} The above four images show a crop of the object we need to describe. 

\textbf{\color{red}Bbox Image Prompt:} The four images above highlight the target object with a red bounding box and dimmed background.

\textbf{\color{red}Task Prompt:} 

\textbf{\color{blue}Caption Task:} Please provide a detailed description of the specified object, focusing on its color, material, shape, state, position, function, surface detail and other information.

(1) Stick to a narrative format for descriptions, avoiding list-like itemizations.

(2) Just output the information you are sure of, if you output the wrong information you will be fired. 

\textbf{\color{blue}Comprehension QA Task:} I need you to generate a series of question pairs for me about this object, using \texttt{<object>} to represent the object in the question pairs. You can focus on its category, color, material, shape, state, position, function, surface detail, size and other information.

"Output example"

Question:  What color is the \texttt{<object>}?

Answer: Mainly red, with some blue as decoration.

Notes:

(1) The object in all images is the same; QA should focus solely on it, without referencing specific images.

(2) Ask as many questions as needed—the more details, the better.

(3) Prioritize reasoning and spatial understanding questions over simple ones.

(4) You can ask questions about the target object by associating it with the surrounding objects (e.g., comparison, spatial relationship, functional relationship, quantitative relationship, etc.). \\

\textit{\# Python code together with above text prompts are directly sent to LLaMA}

\begin{lstlisting}[language=Python, escapeinside={(*@}{@*)}]
messages = [{"role": "user",
    "content": [
    {"type": "text", "text": (*@\textcolor{red}{"Crop Image Prompt"}@*)} + crop_image_list +
      {"type": "text", "text": (*@\textcolor{red}{"Bbox Image Prompt"}@*)} + bbox_image_list +
      {"type": "text", "text": (*@\textcolor{red}{"Task Prompt"}@*)}]}]
\end{lstlisting}

\end{tcolorbox}
\caption{Prompts for object caption and comprehension QA generation. Separate textual instructions are provided for the cropped images and the images highlighting the object via bounding boxes, respectively.}
\label{tab:caption_prompt}
\end{table}


\begin{table}[H]
\centering
\begin{tcolorbox}[colback=gray!10,
                  colframe=black,
                  width=16.5cm,
                  arc=3mm, auto outer arc,
                 ]

\textbf{\color{red}System Prompt:} You are analyzing indoor objects. Given a series of QAs about a single object (marked as \texttt{<object>}), use the information to generate two referring expressions that uniquely identify it.

The two expressions should be:
\begin{itemize}[leftmargin=*,topsep=0pt]
  \item One \textbf{simple referring expression}, using attributes such as category, color, material, spatial location, or function.
  \item One \textbf{situational referring expression}, involving contextual reasoning and diverse sentence structures.
\end{itemize}

\textbf{\color{red}Input Example:}

Question: What is the primary function of the \texttt{<object>}?  
Answer: The \texttt{<object>} is primarily used for holding writing instruments like pens and pencils.


\textit{(Additional QA pairs continue in a similar fashion—omitted for brevity.)}



\textbf{\color{red}Output Example:}

[simple expression]
The cylindrical light brown pen holder on the top shelf of the desk.

[complex expression]
If I finish writing with a pencil, where is the best place to store it?

\end{tcolorbox}
\caption{Prompt for object referring expression generation.}
\label{tab:seg_prompt}
\end{table}

\clearpage

\subsubsection{Spatial QA Generation}
\label{sec:appendix_spatial_qa}
As outlined in Section~\ref{sec:spatial_qa}, we adopt a template-based approach for generating spatial QA. Specifically, we define 14 core spatial abilities and create a total of 30 distinct templates, with each template encompassing at least three different question structures. Some examples of QA templates are provided in Listing~\ref{lst:qa_template}.

\begin{lstlisting}[basicstyle=\ttfamily\small, backgroundcolor = \color{black!10}, caption={Template examples for Spatial QA generation.}, label={lst:qa_template}]
camera_distance_questions = [
    "How far have you walked in total?",
    "What is the total distance you have covered walking?",
    "What is the overall distance you have walked?"
]
closer_to_camera_questions = [
    "Which is closer to you, [A] or [B]?", 
    "Between [A] and [B], which one is nearer to you?",
    "Which one is closer to you, [A] or [B]?"
]
closest_to_camera_questions = [
    "Which is closest to you, [A] or [B] or [C]?", 
    "Among [A], [B], and [C], which one is nearer to you?",
    "Which of [A], [B], or [C] is closest to you?"
]
future_direction_camera_questions = [
    "After you turn 90 degrees to the left, where will [A] be in relation to you?",
    "If you turn left by 90-degree, in which direction will [A] be positioned?",
    "Upon making a 90-degree left turn, how will [A] be oriented with respect to you?"
]
future_direction_camera_rotate_questions = [
    "How many degrees clockwise do you need to turn to face the direction of [A]?",
    "To face towards [A], how many degrees should you rotate in a clockwise direction?",
    "What degree of clockwise rotation is necessary for you to face [A]'s direction?"
]
distance_questions_3 = [
    "Which of the three objects, [A], [B], or [C], is closest to you?",
    "Among [A], [B], and [C], which object is nearest to you?",
    "Between [A], [B], and [C], which one is the closest to you?",
]
height_from_ground_questions = [
    "What is the height difference above ground level between [A] and [B]?",
    "How much higher or lower is [A] compared to [B] above the ground?",
    "By what amount does the elevation of [A] differ from that of [B]?"
]
center_distance_questions = [
    "What is the distance between the centers of [A] and [B]?",
    "How far apart are the centers of [A] and [B]?",
    "What is the separation between the central points of [A] and [B]?"
]
tall_choice_questions_3 = [
    "Among the three objects [A], [B], and [C], which one is the tallest?",
    "Which of the three objects [A], [B], and [C] is tallest?",
    "Out of the three objects [A], [B], and [C], which one is the tallest?",
]
above_predicate_questions = [
    "Is [A] above [B]?",
    "Does [A] appear over [B]?",
    "Can you confirm if [A] is positioned above [B]?",
]
\end{lstlisting}

\subsection{Details of RynnEC-Bench Construction}
\label{sec:rynnec-bench-appe}

As described in Section~\ref{sec:data_balance}, we adjust the object distribution in the object properties understanding evaluation set of RynnEC-Bench based on real-world object category frequencies. The detailed object categorization strategy is presented in Tab.~\ref{tab:object_category}. We classify common indoor objects into 12 coarse categories and 119 fine-grained categories. Objects not falling into any of these predefined categories are assigned to an "other" class. A function-centered taxonomy is adopted: objects with similar appearances but distinct functional roles are categorized separately.

To construct this evaluation set, we follow a two-stage process. First, an initial, oversized pool of 20,000 question-answer (QA) pairs is randomly generated without distributional constraints. Following this, we downsample this pool to a target size of 10,000 pairs. The sampling is performed according to the real-world object distribution outlined previously. Specifically, we calculate the target number of samples for each object category by multiplying its frequency in the distribution by the total target size (10,000). The final dataset is then constructed by drawing the calculated number of QA pairs for each category from the initial 20,000-pair pool. This stratified sampling strategy ensures that the final evaluation set's composition accurately mirrors the specified real-world object frequencies.

\begin{table}[htbp]
\centering
\renewcommand{\arraystretch}{1.3}
\begin{tabularx}{\textwidth}{|l|X|}
\hline
\textbf{Category} & \textbf{Fine-Grained Classes} \\
\hline
Furniture & 
Bed, Chair, Sofa, Table, Nightstand, Cabinet, Shelf, Headboard, Wardrobe, Drawer, Wall, Door, Window, Mirror, Hanger, Hook, Handle, Hinge, Railing, Radiator, Light Switch \\
\hline
Appliances \& Electronics & 
Outlet, Refrigerator, Washing Machine, Air Conditioner, Monitor, Television, Control Panel, Fan, Speaker, Lamp, Charger, Router, Cable, Oven, Toaster, Microwave, Water heater, Range Hoods, Remote Control \\
\hline
Kitchenware \& Tableware & 
Spice Jar, Pot, Kettle, Cup, Jar, Bowl, Spoon, Knife, Plate, Chopping board, Chopstick, Stove, Rice Cooker \\
\hline
Containers & 
Bag, Box, Basket, Bucket, Bottle, Trash Can, Can, Lid, Ashtray \\
\hline
Bathroom \& Cleaning & 
Faucet, Sink, Toilet, Toilet Seat, Toilet Lid, Shower, Bathtub, Mop, Broom, Brush, Sponge, Towel, Toothbrush, Toothpaste, Comb, Soap, Toilet Paper, Hose, Razor, Hair Dryer \\
\hline
Textiles \& Bedding & 
Quilt, Blanket, Carpet, Curtain, Pillow, Cushion, Mattress \\
\hline
Stationery \& Office Supplies & 
Book, Clock, Calendar, Pen, Sharpener, Scissors, Calculator, Mouse, Mousepad, Keyboard, LaptopPanel, Tablet Computer \\
\hline
Decor \& Art & 
Plant, Painting, Picture, Poster, Label, Calendar, Vase \\
\hline
Daily Necessities & 
Phone, Hat, Slipper, Shoe, Umbrella, Headphones, Glove \\
\hline
Food & 
Fruit, Vegetable \\
\hline
Clothing & 
Shirt, Pants, Dress, Skirt, Coat, Shorts, Socks, Underwear \\
\hline
Fitness \& Recreation & 
Treadmill, Dumbbells, Piano, Toy \\
\hline
\end{tabularx}
\caption{Object category taxonomy.}
\label{tab:object_category}
\end{table}

\subsection{Qualitative examples}
In \cref{main-result} and \cref{tab:main_result}, we show that our model can handle different types of object and spatial cognition tasks. In this section, we show more detailed qualitative examples for different abilities of our model.

\subsubsection{Object Cognition}
\begin{itemize}
\item \textbf{Properties (\cref{fig:qual-1}, \cref{fig:qual-2}).} The model discerns a wide range of object properties, including physical attributes such as size, color, and surface details, as well as functional affordances.
\item \textbf{Segmentation (\cref{fig:qual-1}).} The system performs both simple and situational referring expression segmentation, enabling it to isolate target objects in the scene based on natural language queries.
\end{itemize}

\subsubsection{Spatial Cognition}
\begin{figure*}[t]
    \centering
    \includegraphics[width=\textwidth]{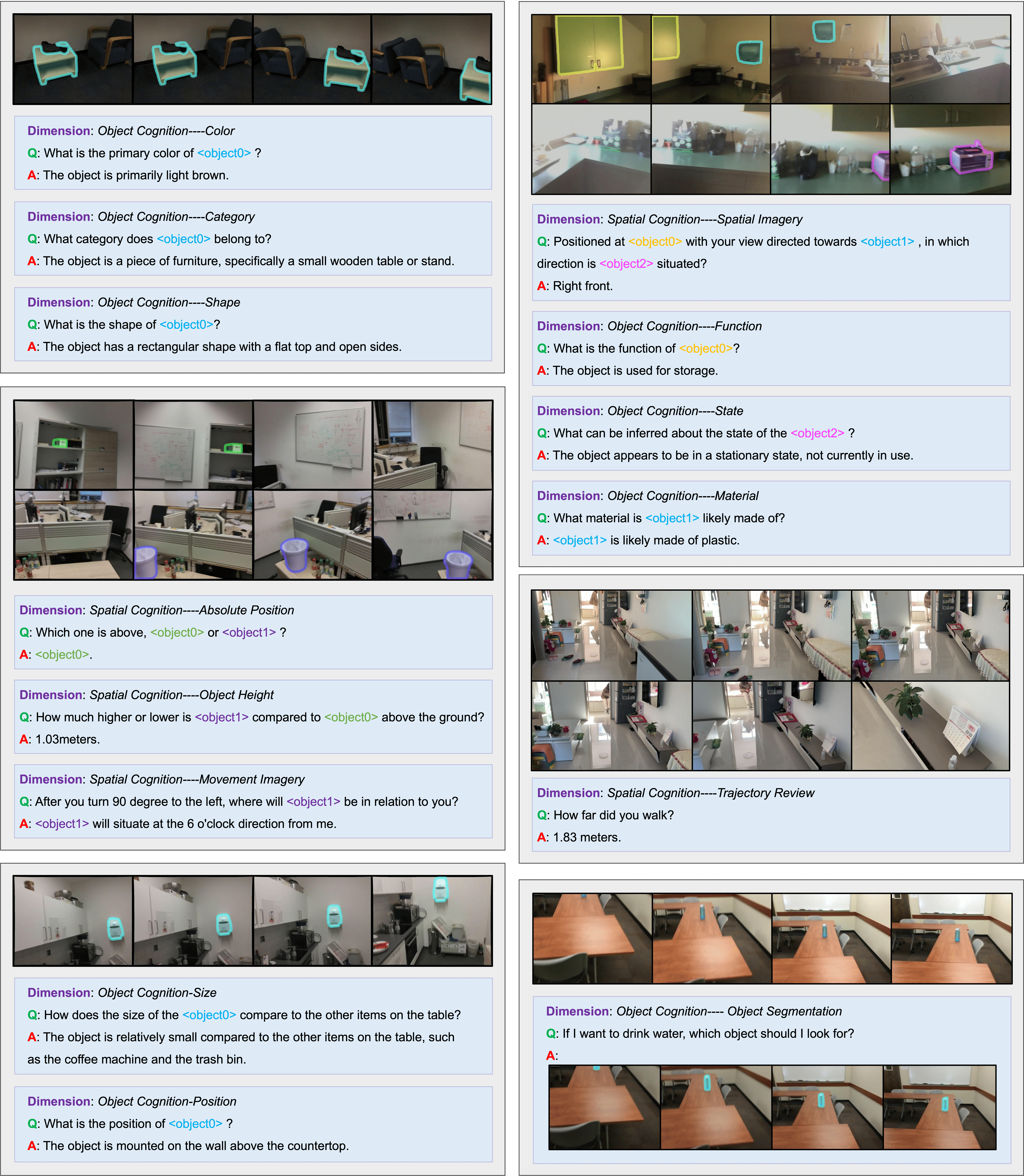}

    \caption{\textbf{Visualization of question answering examples.} Part
1 out of 2.}
    \label{fig:qual-1}
\end{figure*}

\begin{figure*}[t]
    \centering
    \includegraphics[width=\textwidth]{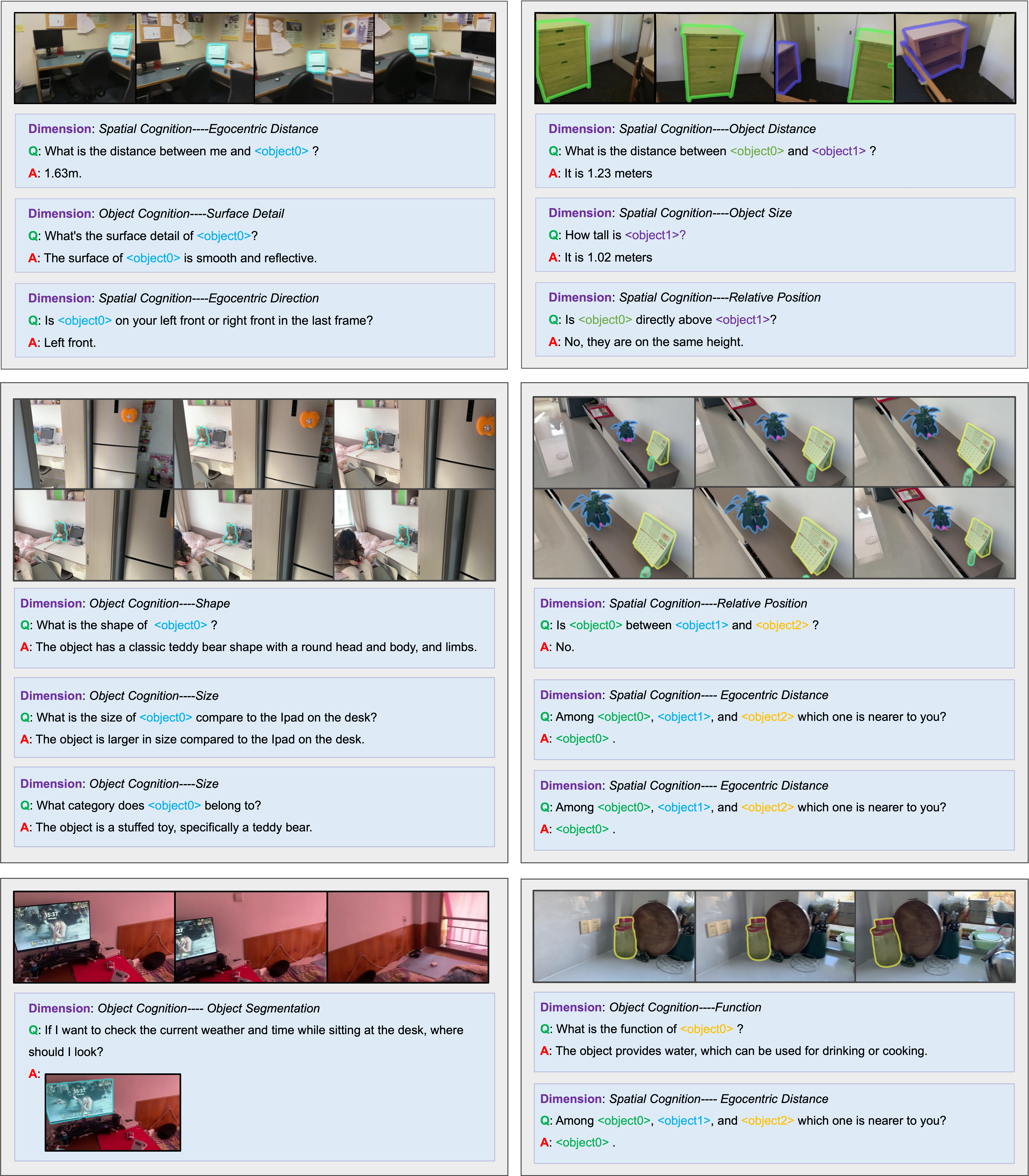}

    \caption{\textbf{Visualization of question answering examples.} Part
2 out of 2.}
    \label{fig:qual-2}
\end{figure*}

\begin{itemize}
\item \textbf{Trajectory Review (\cref{fig:qual-1}).} The model perceives the distance traversed by its own camera, allowing for a review of its past trajectory.
\item \textbf{Egocentric Direction (\cref{fig:qual-2}).} It successfully determines the direction of objects relative to its own perspective.
\item \textbf{Egocentric Distance (\cref{fig:qual-2}).} The system is capable of estimating the egocentric distance from itself to surrounding objects in the environment.
\item \textbf{Movement Imagery (\cref{fig:qual-1}).} A key capability is the imagination of prospective movements, allowing the model to reason about future paths.
\item \textbf{Spatial Imagery (\cref{fig:qual-1}).} The model demonstrates an ability for spatial imagination, such as inferring the layout of unseen areas.
\item \textbf{Object Size (\cref{fig:qual-2}).} Its spatial understanding extends to estimating the absolute sizes of objects and performing relative size comparisons between them.
\item \textbf{Object Height (\cref{fig:qual-1}).} Similarly, the model predicts and compares the heights of different objects.
\item \textbf{Object Distance (\cref{fig:qual-2}).} The system accurately gauges the distance between multiple objects within the scene (i.e., inter-object distance).
\item \textbf{Absolute Position (\cref{fig:qual-1}).} The model can ascertain the absolute positional relationships between objects.
\item \textbf{Relative Position (\cref{fig:qual-2}).} Furthermore, it demonstrates a robust understanding of the relative positions of objects with respect to one another.
\end{itemize}

\end{document}